\documentclass[11pt]{article}

\usepackage{acl}

\usepackage{times}
\usepackage{latexsym}

\usepackage[T1]{fontenc}

\usepackage[utf8]{inputenc}

\usepackage{microtype}

\usepackage{inconsolata}

\usepackage{enumitem} 
\usepackage{graphicx}
\usepackage{makecell}
\usepackage{tabularx}
\usepackage{booktabs}
\usepackage{amsmath}
\usepackage{amssymb}
\usepackage{cleveref}
\usepackage{multirow}
\crefname{section}{Section}{Sections}  

\usepackage{tcolorbox}
\usepackage{array}
\usepackage{xcolor}
\usepackage{diagbox}

\usepackage{subcaption} 

\usepackage{todonotes}
\usepackage[normalem]{ulem}
\usepackage{xspace}
\usepackage{float} 

\makeatletter
\newcommand*\iftodonotes{\if@todonotes@disabled\expandafter\@secondoftwo\else\expandafter\@firstoftwo\fi}  
\makeatother

\title{Evaluating Communicative Belief Updates in Large Language Models\linebreak via Implicature Recognition and Cancellation}

\usepackage{xspace}
\newcommand{\datasetname}{\textsc{ImplicatureX}\xspace}

\newcommand{\negdataset}{\textsc{Implicature}$^\bot$\xspace}
\newcommand{\neutraldataset}{\textsc{Implicature$^\approx$}\xspace}
\newcommand{\strengthdataset}{\textsc{Implicature$^+$}\xspace}

\newcommand{\scalarimpli}[2]{$\langle$\textit{#1}, \textit{#2}$\rangle$\xspace}
\newcommand{\someall}{$\langle$\emph{some}, \emph{all}$\rangle$\xspace}

\definecolor{contextcol}{RGB}{0, 158, 115}
\definecolor{trigcol}{RGB}{204, 121, 167}
\definecolor{cancelcol}{RGB}{230, 159, 0}
\definecolor{implicaturecol}{RGB}{0, 114, 178}
\definecolor{cgcol}{RGB}{0, 0, 0}

\newcommand{\contextcolor}[1]{\textcolor{contextcol}{#1}}
\newcommand{\trigcolor}[1]{\textcolor{trigcol}{#1}}
\newcommand{\cancelcolor}[1]{\textcolor{cancelcol}{#1}}
\newcommand{\implicature}[1]{\textcolor{implicaturecol}{#1}}
\newcommand{\cgcolor}[1]{\textcolor{cgcol}{#1}}

\newcommand{\context}{\contextcolor{c}}
\newcommand{\trig}{\trigcolor{u}}
\newcommand{\cancel}{\cancelcolor{u^\times}}
\newcommand{\strength}{\cancelcolor{u^+}}
\newcommand{\neutral}{\cancelcolor{u^\approx}}
\newcommand{\explicit}{\cancelcolor{u^\bot}}
\newcommand{\belief}{\implicature{b}}
\newcommand{\beliefspace}{\implicature{B}}
\newcommand{\ctxspace}{\contextcolor{\mathcal{C}}}
\newcommand{\uttspace}{\trigcolor{\mathcal{U}}}      
\newcommand{\uttseq}{\trigcolor{\mathcal{U}^*}}      
\newcommand{\cg}{\cgcolor{G}}

\newcommand{\cgfunc}{\ensuremath{\cg : \mathcal{\beliefspace} \times \ctxspace \times \uttseq \rightarrow [0,1]}}
\newcommand{\cgprior}{\ensuremath{\cg(\belief \mid \emptyset, \emptyset)}}

\newcommand{\utthistory}{\trigcolor{\mathbf{u}}}
\newcommand{\cgadd}[2]{\ensuremath{\cg(\implicature{#1} \mid \context, #2) > \cg(\neg\implicature{#1} \mid \context, #2)}}
\newcommand{\cgrem}[3]{\ensuremath{\cg(\implicature{#1} \mid \context, #2) < \cg(\implicature{#1} \mid \context, #3)}}

\newcommand{\probimp}{\ensuremath{P_{\mathcal{M}}(\belief \mid \mathrm{\texttt{t}}_{\belief}(\context, \langle \trig \rangle))}}
\newcommand{\probnegimp}{\ensuremath{P_{\mathcal{M}}(\neg\belief \mid \mathrm{\texttt{t}}_{\belief}(\context, \langle \trig \rangle))}}
\newcommand{\probcancel}{\ensuremath{P_{\mathcal{M}}(\belief \mid \mathrm{\texttt{t}}_{\belief}(\context, \langle \trig, \cancel \rangle))}}

\newcommand{\probstrength}{\ensuremath{P_{\mathcal{M}}(\belief \mid \mathrm{\texttt{t}}_{\belief}(\context, \langle \trig, \strength \rangle))}}
\newcommand{\probneutral}{\ensuremath{P_{\mathcal{M}}(\belief \mid \mathrm{\texttt{t}}_{\belief}(\context, \langle \trig, \neutral \rangle))}}
\newcommand{\probnegneutral}{\ensuremath{P_{\mathcal{M}}(\neg \belief \mid \mathrm{\texttt{t}}_{\belief}(\context, \langle \trig, \neutral \rangle))}}

\newcommand{\beliefsym}{b}
\newcommand{\contextsym}{c}
\newcommand{\trigsym}{u}
\newcommand{\cancelsym}{u^\times}
\newcommand{\coloredstatement}[3]{%
  #2{\ensuremath{#1 =}\ ``#3''}\xspace%
}
\newcommand{\beliefstmt}[1]{\coloredstatement{\beliefsym}{\implicature}{#1}}
\newcommand{\contextstmt}[1]{\coloredstatement{\contextsym}{\contextcolor}{#1}}
\newcommand{\trigstmt}[1]{\coloredstatement{\trigsym}{\trigcolor}{#1}}
\newcommand{\cancelstmt}[1]{\coloredstatement{\cancelsym}{\cancelcolor}{#1}}

\usepackage{tikz}
\definecolor{implicGreen}{HTML}{2D5A43}
\definecolor{implicRed}{HTML}{8B3A3A}
\definecolor{implicBlue}{HTML}{205090}
\definecolor{blueSynthetic}{HTML}{3B71B8} 
\definecolor{blueNatural}{HTML}{153560} 
\newcommand*\circled[2]{%
  \begin{tikzpicture}[baseline=(char.base)]
    \node[shape=circle, draw=#1, fill=#1, inner sep=1.2pt, text=white, font=\small\bfseries] (char) {#2};
  \end{tikzpicture}%
}

\usepackage{fontawesome5}  

\author{
  \textbf{Cesare Spinoso-Di Piano}$^{1}$ 
  \textbf{Verna Dankers}$^{1\,\text{\faCoffee}}$ 
  \textbf{Marius Mosbach}$^{1\,\text{\faCoffee}}$ 
  \textbf{Jackie Chi Kit Cheung}$^{1,2}$ \\ 
  \textnormal{$^1$Mila - Quebec AI Institute \& McGill University, $^2$Canada CIFAR AI Chair} \\
  \texttt{\{cesare.spinoso, cheungja\}@mila.quebec} 
}

\begin{document}

\maketitle

\begin{abstract}
Human language is driven by unspoken beliefs and belief updates, making these critical to model for successful communication between large language models (LLMs) and their users. In this paper, we evaluate the ability of LLMs to recognize unspoken beliefs made through \emph{implicatures} and to understand their updates through \emph{implicature cancellation}: the pragmatic phenomenon whereby an utterance's implied meaning is \emph{weakened} or \emph{negated}. We create the first expert-annotated implicature cancellation dataset, \datasetname, crowdsourced for human judgements of implicatures and their corresponding cancellations. We find that LLM belief update understanding lags behind that of humans, especially in more naturally-occurring scenarios. Additional control experiments suggest that successes in LLM belief updates may stem in part from a reliance on prior beliefs, and that failures in belief updates may depend on their type and on their form. Overall, our study suggests that current LLMs have not yet reached human-level understanding of unspoken beliefs and belief updates.\footnote{Code and data available at \url{https://github.com/cesare-spinoso/ImplicatureX}.}
\end{abstract}

\def\thefootnote{\faCoffee}\footnotetext{~Equal contribution.}\def\thefootnote{\arabic{footnote}}

\section{Introduction}
\label{sec:introduction}

\begin{figure}[t]
    \centering
    \includegraphics[width=\linewidth]{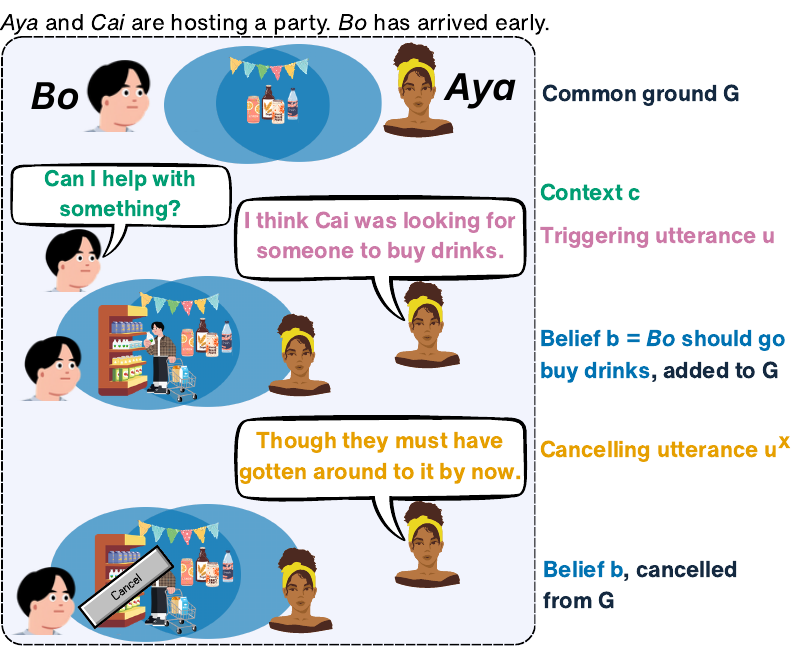}
    \caption{An example of belief negotiation: A belief $\belief$ in the common ground $\cg$ is updated as an implicature is triggered by $\trig$ and then cancelled by $\cancel$.
    }
    \label{fig:figure1}
\end{figure}

Natural language communication consists of threads of beliefs negotiated between interlocutors in a conversational common ground \citep{lewis1979scorekeeping,heim1982semantics,clark1991brennan,stalnaker1998representation,kamp2013theory}.
These beliefs are often introduced and updated implicitly through \emph{implicatures} and \emph{implicature cancellations}, whereby a previously implicated belief is \emph{negated} or \emph{weakened}  \citep{grice1975logic,sperber1986relevance}. 
For instance, when Bo asks their friend Aya whether they can help with something, the response ``I think Cai was looking for someone to buy drinks.'' triggers an implicature that Cai needs Bo to buy drinks (Figure~\ref{fig:figure1}). However, Aya can cancel this implicature by saying ``Though they must have gotten around to it by now.'', at which point Bo should update their beliefs and act accordingly, e.g., by asking Cai to confirm.
Interlocutors' beliefs can thus change from one utterance to the next. Understanding these beliefs and how they are updated is thus of fundamental importance for successful interactions between large language models (LLMs) and human system users.

In this work, we evaluate LLMs' ability to understand such dynamically changing beliefs in language. 
We focus on implicature recognition and cancellation, two of the most widely studied and fundamental phenomena involving belief updates in linguistics and cognitive science 
\citep{grice1975logic,sperber1986relevance}. 
While previous studies have evaluated the ability of LLMs to perform belief updates of world knowledge and logical reasoning \citep{rudinger2020thinking,hwang2021comet}, similar evaluations in the communicative space remain underexplored. 
Our work fills this gap by evaluating LLMs' ability to identify beliefs triggered by implicatures and to update those implicated beliefs as a result of implicature cancellation.

We create the first implicature cancellation dataset, \datasetname, consisting of $271$ expert-annotated implicatures and corresponding cancellations. 
Beyond synthetic two-turn conversational implicatures, it contains naturally occurring scalar implicatures, discourse implicatures and multi-turn dialogue conversational implicatures. In addition, we conduct and include a crowdsourcing annotation of our \datasetname items to measure implicature and cancellation recognition accuracies.

Our results show that LLMs' pragmatic reasoning falls short of a human understanding of unspoken beliefs conveyed by implicatures. While the strongest LLMs we tested (e.g., GPT-5.4 Thinking) match human performance on implicature recognition of scalar and discourse implicatures, LLMs struggle to perform at a level above random chance on naturally-occurring conversational implicatures. 
Moreover, a control experiment reveals that, in many cases, LLMs successfully recognize implicatures without observing the context or the utterance. We, therefore, question the extent to which their success is due to legitimate pragmatic reasoning.

In a similar vein to our implicature recognition findings, we find that belief updates following implicature cancellation---which are readily made by humans---are especially challenging for LLMs in naturally-occurring and realistic scenarios.
Additional control experiments reveal that the type of belief update---cancelling, leaving unchanged, or strengthening---and the way in which the update is triggered---explicitly or implicitly---affect the extent to which LLMs are able to revise their existing beliefs. For instance, our results demonstrate that even when presented with explicit belief negations, strong LLMs (e.g., Qwen 3 32B Thinking) cannot match human accuracy in cancellation recognition.

To summarize, we study the extent to which LLMs are able to understand unspoken beliefs via implicature recognition and belief updates via implicature cancellation, two fundamental properties of human communication. We create the first implicature cancellation dataset, \datasetname, verified by linguistic experts and annotated with crowdsourced judgements. Our evaluation reveals that LLMs fall short of both a human understanding of unspoken beliefs and of belief updates. More broadly, our work suggests that LLMs may still not be able to understand the nuances of belief productions and negotiations, especially when they are unspoken and naturally-occurring.

\section{Related Work}
\label{sec:related_work}

\paragraph{Philosophy of language}

Communication has long been thought to be an exercise of \emph{belief negotiation} \citep{grice1957meaning, lewis1979scorekeeping}. 
In this vein, several seminal studies have posited that we communicate by making updates to an ever-evolving tacitly agreed upon set of propositions, i.e., a \emph{common ground} \citep{heim1982semantics,clark1991brennan,stalnaker1998representation,kamp2013theory}. 
As such, implicatures---beliefs about a possible intended meaning suggested by a speaker---and implicature negotiations---the cancellation and strengthening of these beliefs---are a core mechanism by which we add and make updates to the common ground \citep{grice1975logic}. 
In this work, we offer an experimental account of belief updates through implicature cancellations which, unlike implicature strengthenings \citep{benotti2010implicature}, remain an understudied area of belief negotiation.

\paragraph{Experimental pragmatics}

Belief negotiation in human communication has received widespread attention in experimental pragmatics. 
Initially studied in the context of referring expressions \citep{clark1986referring, isaacs1987references,selten2007emergence,deemter2012generation}, negotiations of common ground beliefs have continued to be of central relevance to other pragmatic phenomena including non-verbal communication \citep{veinott1999video,clark2004speaking}, discourse relations \citep{fetzer2018linguistic} and scalar implicatures \citep{noveck2001children}. 
In particular, studies have shown that the inferences made from scalar implicatures are more easily and more readily accepted into the conversational common ground based on the context in which they appear and the prior beliefs held by interlocutors \citep{breheny2006generalised,grodner2010some,degen2015investigating,yang2018context,huang2018some}. 
Our study provides a continued experimental investigation of communicative belief updates through the phenomenon of implicature cancellation.

\paragraph{Communicative beliefs in NLP}

Notions surrounding communicative beliefs have long served as inspiration for language generation systems, including dialogue systems \citep{allen1980analyzing,grosz1986attention,dale1995computational}, image captioning \citep{andreas2016reasoning}, and conversational agents \citep{korner2025common}. 
Furthermore, implicatures and unspoken beliefs have been used extensively to evaluate the communicative competence of LLMs \citep{jeretic2020natural,kabbara-cheung-2022-investigating,ruis2023goldilocks,cho2024pragmatic,yue2024large,cong2024manner}. 
Our contribution differs in that we leverage the cancellability of implicatures as a tractable space to evaluate the ability of LLMs to \emph{update} unspoken beliefs. 
Finally, while processes similar to implicature cancellation, such as abductive and defeasible reasoning, have been studied extensively \citep{lascarides1991discourse,rudinger2020thinking,hwang2021comet}, they differ from our focus of communicative belief updates.

\section{Operationalizing Implicature and Implicature Cancellation}
\label{sec:theoretical_setup}

We view any communicative exchange (spoken, written, signed, etc.) as consisting of a
sequence of \emph{utterances} $\trig \in \uttspace$, i.e., any unit of language produced
by a participant of the exchange.
Further, let $\uttseq$ denote the set of all possible utterance sequences and $\utthistory = \langle \trig_1, \ldots, \trig_n \rangle \in \uttseq$ a communicative history.
Each exchange takes place in a \emph{context} $\context \in \ctxspace$, which captures situational and
communicative information relevant to interpreting the utterances.\footnote{$\context$ is assumed to be fixed for the duration of the exchange.}\looseness-1

Participants share a set of \emph{beliefs} $\mathcal{\beliefspace}$, i.e., propositional statements such as ``Cai needs help buying drinks.'', which represent what they mutually take to be true.
Formally, the \emph{common ground} is a function $\cgfunc$ that, given $\context$ and $\utthistory$, assigns to each $\belief \in \mathcal{\beliefspace}$ a
probability reflecting how strongly that belief is mutually held by the
participants.\footnote{Prior to any exchange, the common ground is given by $\cgprior$,
which may assign uniform probability to all $\belief \in \mathcal{\beliefspace}$, or
reflect presupposed propositions based on the participants' shared history and prior beliefs \citep{anderson2018essentials}.}

A belief $\belief$ becomes part of the common ground via an \emph{implicature} when
utterance $\trig$, produced in context $\context$, makes $\belief$ more probable than its
negation according to the common ground, i.e., $\cgadd{b}{\langle \trig \rangle}$.
A belief $\belief$ is subsequently \emph{cancelled} from the common ground via an \emph{implicature cancellation} when a speaker extends the communicative history from $\langle\ldots,
\trig\rangle$ to $\langle\ldots, \trig, \cancel\rangle$ with a cancelling utterance
$\cancel$, which weakens or negates the implicature triggered by $\trig$, i.e.,
$\cgrem{b}{\langle \trig, \cancel \rangle}{\langle \trig \rangle}$.
Thus, we define a \emph{belief update} in the context of implicature cancellation as $\cgrem{b}{\langle \trig, \cancel \rangle}{\langle \trig \rangle}$ when provided with $\cancel$ for an utterance $\trig$ for which $\cgadd{b}{\langle \trig \rangle}$ holds.

\begin{figure*}[t]
    \centering
    \includegraphics[width=\linewidth]{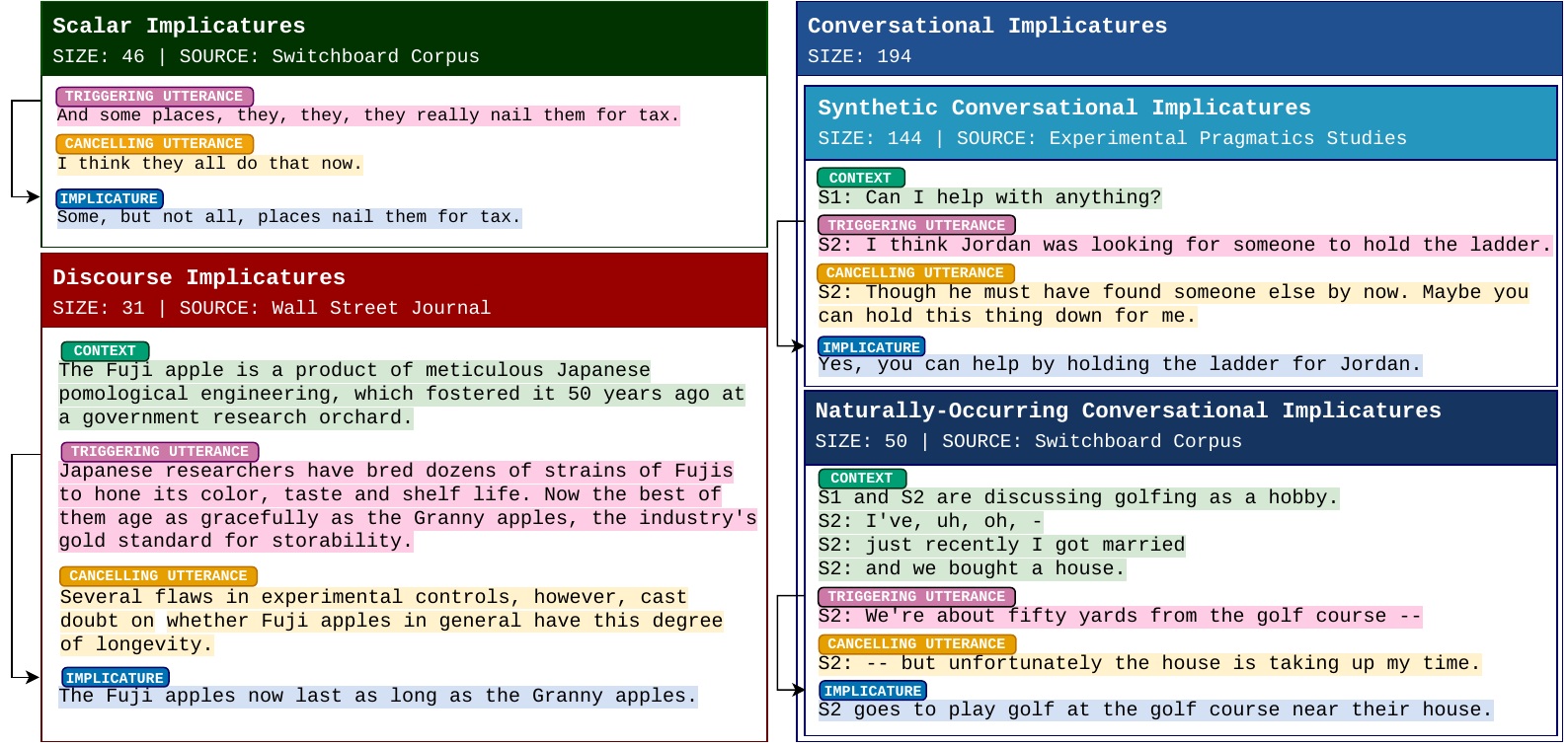}
    \caption{Composition of the \datasetname dataset with sizes, sources and illustrative examples. Note that the sizes reported in this figure reflect the dataset post cleaning (\Cref{sec:expert_annotation}).
    }
    \label{fig:dataset_composition}
\end{figure*}

\section{The \datasetname Dataset}
\label{sec:dataset}

Here, we describe the composition of the \datasetname implicature cancellation dataset along with its expert and crowdsourcing annotation.

\subsection{Dataset Composition}

\datasetname is an expert-annotated dataset of $271$ items which include \circled{implicGreen}{1} \textsc{Scalar Implicatures}, \circled{implicRed}{2} \textsc{Discourse Implicatures} and \circled{implicBlue}{3} \textsc{Conversational Implicatures}.
\Cref{fig:dataset_composition} provides example stimuli per implicature type.
Below, we describe the dataset's composition.

\circled{implicGreen}{1} \textsc{Scalar Implicatures.} 
Our items are based on the well-known pragmatic phenomenon where, given a pair of lexical items \scalarimpli{$w_1$}{$w_2$} with $w_2$ semantically entailing $w_1$, the use of $w_1$ in an utterance indicates the negation of $w_2$. 
For instance, the triggering utterance \trigstmt{\emph{Some} places nail them for tax.} increases the probability of the belief \beliefstmt{\emph{Not all} places nail them for tax.}
belonging to the common ground. 
This belief can be revised by the speaker using a cancelling utterance \cancelstmt{I think they \emph{all} do that now.}.
We sample 50 naturally-occurring \someall scalar implicatures from the Switchboard Corpus \citep{switchboard_corpus} as originally collected by \citet{degen2015investigating} and manually add an implicature cancellation to each sampled item.

\circled{implicRed}{2} \textsc{Discourse Implicatures.} We posit that discourse relations, especially those which involve implicit causal relations, may be recast as \emph{discourse} implicatures---i.e., implicatures found in traditional forms of discourse. 
In particular, we identify and extract discourse implicatures from Wall Street Journal (WSJ) articles, using the corresponding implicit discourse relations annotated in the Penn Discourse Tree Bank (PDTB) corpus \citep{webber2019penn}. Using the implicit PDTB causal discourse relations, we are able to identify excerpts such as \contextstmt{The Fuji apple has been extensively researched.} and \trigstmt{Strains have been developed to age as gracefully as the Granny apples.} which implicate the belief \beliefstmt{Fuji apples last as long as Granny apples.} This belief is negated---or at least weakened---with the cancelling utterance \cancelstmt{Though flaws in experimental controls now cast doubt on the apple's longevity.} 
We identify and annotate 31 WSJ article excerpts using implicit causal relations from the PDTB corpus.

\circled{implicBlue}{3} \textsc{Conversational Implicatures.} Our conversational implicatures consist of implicatures which are triggered by utterances produced in a transcribed exchange between two conversational participants. 
These 197 conversational implicatures are further subdivided into two categories which differ principally by their source: \circled{blueSynthetic}{A} Synthetic conversational implicatures and \circled{blueNatural}{B} Naturally-occurring conversational implicatures.%

\circled{blueSynthetic}{A} The synthetic conversational implicatures consist of an exchange between two participants, $S_1$ and $S_2$. 
The exchange begins typically with a question (\contextstmt{Do you need help with anything?}) which is followed by a response by the other participant (\trigstmt{I think $C$ was looking for someone to do $X$.}), introducing an implicated belief into the common ground (\beliefstmt{You can help by doing $X$ for $C$.}). 
This belief is updated via a cancelling utterance (\cancelstmt{Though they must be done by now.}). 
We collect 146 such two-turn conversational implicatures from several existing implicature understanding datasets 
\citep{george2020conversational,louis2020d,wilson2021second,hu2023fine} and manually add implicature cancellations.\looseness-1   

\circled{blueNatural}{B} Our scenario-based conversational implicatures also consist of an exchange between $S_1$ and $S_2$. 
In this case, the exchange is a naturally-occurring discussion between two Switchboard Corpus participants \citep{switchboard_corpus}. 
We collect 51 naturally-occurring conversational implicatures. 
The cancelling utterances are produced by one of the participants within the exchange. For instance, \trigstmt{We're close to the golf course.} (\beliefstmt{They go golfing near their home}) is cancelled by \cancelstmt{Unfortunately the house is taking up time.}
    
\subsection{Expert Annotation}

\label{sec:expert_annotation}

\paragraph{Annotation details} 

To validate the plausibility of the implicatures and the correctness of their cancellations in \datasetname, we ran an expert annotation of our implicature cancellation items. We elicit expert judgements for the implicatures' plausibility as well as the cancellations' correctness. In addition, we elicit alternatives for implausible implicatures or incorrect cancellations (where possible).

The stimuli were subdivided into six batches of 53 items each of which included seven attention checks, i.e., items with either a clearly implausible implicature or incorrect cancellation. 
The seven attention checks were different for each batch and their frequency per batch reflected the makeup of the dataset: one scalar implicature, one discourse implicature, four synthetic conversational implicatures and one naturally-occurring conversational implicature.

We hired two professionally trained linguists for two rounds of expert annotation. In the first annotation round, both expert annotators were given the same batch in order to compute inter-annotator agreement of implicature plausibility and cancellation correctness. In the second round of annotations, one of the expert annotators was tasked to annotate the five remaining batches.

Additional details regarding the expert annotation are presented in \Cref{sec:app:expert_annotation}.

\paragraph{Annotation results} 

For the first round of annotation, we report the raw agreement, Cohen's kappa ($\kappa$) and the prevalence-adjusted bias-adjusted kappa (PABAK) for implicature plausibility and cancellation correctness\footnote{To compute this second agreement, we exclude items where at least one of the annotators marked the implicature as implausible.} in Table~\ref{tab:agreement}. The raw agreement and PABAK are relatively high. The relatively low Cohen's $\kappa$ is explained by the imbalance in class distributions for both the implicature plausibility and cancellation correctness annotation \citep{byrt1993bias} (See confusion matrices in Tables~\ref{tab:confusion-plausibility}~and~\ref{tab:confusion-cancellation} in Appendix~\ref{sec:app:expert_annotation}).

\begin{table}[htbp]
    \centering
    \resizebox{\columnwidth}{!}{%
    \begin{tabular}{lccc}
        \toprule
        & \makecell{Raw\\Agreement} & Cohen's $\kappa$ & PABAK \\
        \midrule
        \makecell[l]{Implicature\\Plausibility}  & 0.85 & 0.34 & 0.70 \\
        \makecell[l]{Cancellation\\Correctness}  & 0.86 & 0.32 & 0.72 \\
        \bottomrule
    \end{tabular}}
    \caption{Inter-annotator agreement scores per annotation task.}
    \label{tab:agreement}
\end{table}

The second round of annotation generated 18 implicature replacements and 27 cancellation replacements. In seven cases, the items were flagged but the annotator was not able to provide alternatives (e.g., the cancellations for the naturally-occurring conversational implicatures). 
After dropping and replacing the flagged items, this left us with a total of 271 expert-annotated implicature cancellation items which are distributed as follows: \circled{implicGreen}{1}: $46$ items, \circled{implicRed}{2}: $31$ items, \circled{implicBlue}{3}\circled{blueSynthetic}{A}: $144$ items, \circled{implicBlue}{3}\circled{blueNatural}{B}: $50$ items.

\subsection{Crowdsourcing Annotation}
\label{sec:main:crowdsourcing_annotation}

While our expert annotation served to validate the plausibility and correctness of \datasetname, we perform a crowdsourcing annotation to provide a realistic topline comparison to LLM understanding of implicature and implicature cancellation.

\paragraph{Annotation details} 

To obtain aggregated human judgments of \datasetname belief updates, we run a crowdsourcing annotation of the $271$ expert annotated items. To do so, we elicit likelihood judgements for each implicature item given its corresponding context, triggering utterance, and cancelling utterance. In particular, we create $16$ stimuli batches of $30$ items and two stimuli batches of $32$ items. For each batch, half of the items contain the cancelling utterance and half do not. We shuffle batch items in random order and ensure that for every batch no overlap exists between the items with and without the cancelling utterance. In addition, we include and reuse $10$ attention checks for every batch, five whose likelihood rating should be high and five whose likelihood rating should be low.

To implement our likelihood judgement elicitation, we generalize the approach from experimental pragmatics for studying the strength of scalar implicatures to our entire set of stimuli. In particular, we generalize \citet{degen2015investigating}'s approach and ask participants to rate the likelihood of the implicature on a seven point Likert scale with endpoints labeled as ``absolutely impossible'' and ``absolutely certain'' and individual points labeled as 1, 2, \dots, 7.

We recruit $90$ participants from the Prolific platform based in Canada, the U.S. and the U.K. whose self-reported native language is English. In addition, we select participants with an undergraduate degree and who have taken part in at least 100 studies with a hit rate greater than or equal to $99/100$. Each batch is annotated by 5 participants. We exclude a participant's responses if they fail 3 or more attention checks.

Additional details regarding the crowdsourcing annotation are presented in \Cref{sec:app:crowdsourcing_annotation}.

\begin{figure}[t]
    \centering
    \includegraphics[width=\linewidth]{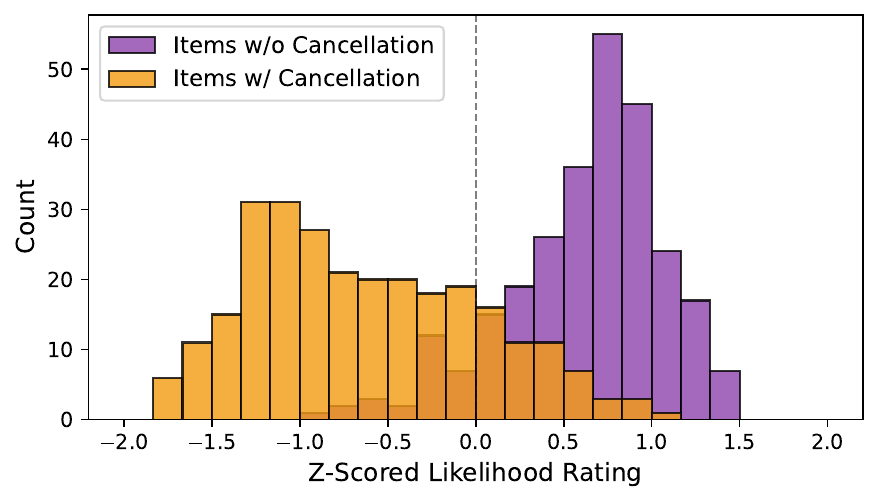}
    \caption{Histogram of average likelihood judgements z-scores from \datasetname with and without the cancelling utterance.
    }
    \label{fig:main:human_ratings}
\end{figure}

\paragraph{Annotation results}

Since human annotators are known to interpret Likert scales differently \citep{cliff1993dominance}, we apply per-participant z-scoring to the Likert scale likelihood judgements. We present z-scored likelihood judgements averaged per item in \Cref{fig:main:human_ratings}.
\Cref{fig:main:human_ratings} shows a clear belief update given cancelling utterances: items without a cancelling utterance tend to have a positive z-score while items with a cancelling utterance tend to have a negative z-score. Additional results related to our crowdsourcing annotation can be found in \Cref{sec:app:crowdsourcing_annotation}.

\begin{figure*}[t]
    \centering
    \includegraphics[width=\linewidth]{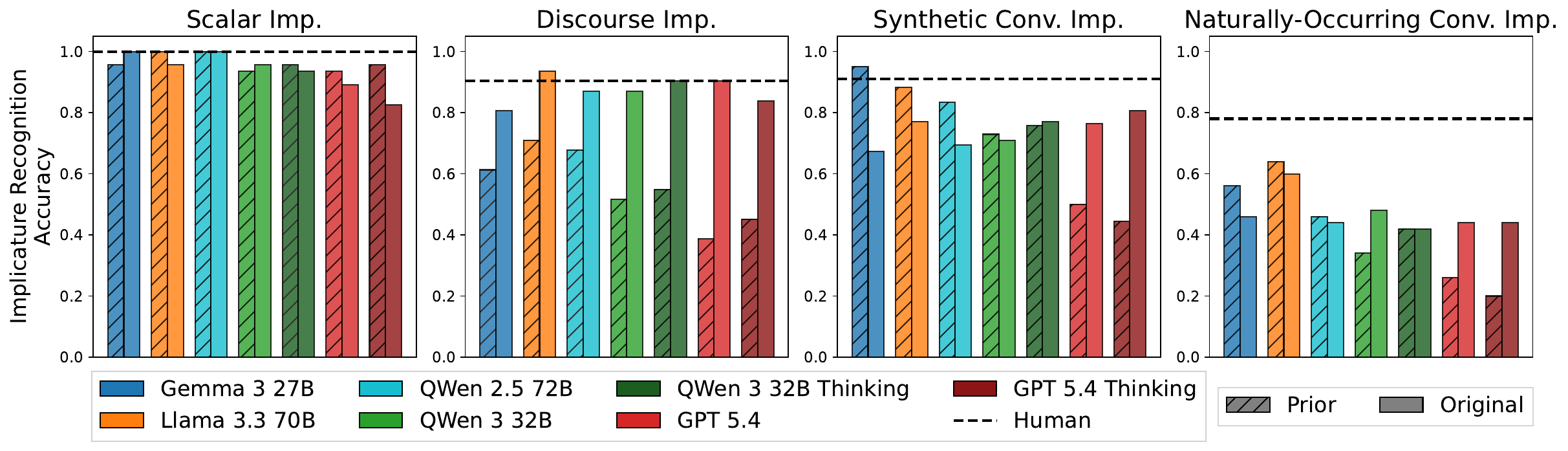}
    \caption{Implicature recognition accuracies of the largest and most modern models for each class of models tested. Original denotes the accuracy using the $\mathrm{\texttt{t}}_{\belief}(\context,
\utthistory))$ prompt and prior denotes the accuracy using the $\mathrm{\texttt{t}}_{\belief}(\emptyset, \emptyset))$ prompt.
    }
    \label{fig:main:impli_prior_contol}
\end{figure*}
\section{Task Definitions and Setup}

We leverage the \datasetname dataset to evaluate LLM understanding of implicature, implicature cancellation, and of belief updates induced by these two pragmatic phenomena. To do so, we use a multiple-choice-question prompt template to estimate specific values
of the common ground function $\cgfunc$.
Given a belief $\belief$ and a conversational history $\utthistory \in \uttseq$, we
operationalize $\cg(\belief \mid \context, \utthistory)$ via an LLM $\mathcal{M}$'s
next-token probability distribution over its token space $\mathcal{V}$, such that: $\cg(\belief \mid \context, \utthistory) \approx P_{\mathcal{M}}(\belief \mid \mathrm{\texttt{t}}_{\belief}(\context, \utthistory)).$
Here, $\mathrm{\texttt{t}}_{\belief}(\context, \utthistory)$ is a prompt containing the
verbalization of $\belief$, the context $\context$, the conversational history
$\utthistory$, and a question about the truthfulness of $\belief$ in a multiple-choice
format using ``True'' and ``False'' as options (Example prompt in \Cref{fig:prompt-template,fig:prompt-example} in \Cref{sec:app:prompt_templates}).
To reduce positional bias, we follow \citet{shi-etal-2025-judging} and
shuffle the order of the option choices, creating two copies for each prompt.

\paragraph{Implicature recognition}

We evaluate an LLM $\mathcal{M}$'s \textit{implicature recognition accuracy} by computing the proportion of items for which $\mathcal{M}$ favors the implicated belief over its negation, i.e., 
\begin{align*}
    P_{\mathcal{M}}(\belief \mid \mathrm{\texttt{t}}_{\belief}(\context, \utthistory)) &> 
    P_{\mathcal{M}}(\neg \belief \mid \mathrm{\texttt{t}}_{\belief}(\context, \utthistory)).
\end{align*}

\paragraph{Cancellation recognition}

We compute $\mathcal{M}$'s \textit{cancellation recognition accuracy} as the proportion 
of items for which $\mathcal{M}$ decreases its probability of $\belief$ upon observing $\cancel$, i.e.,
\begin{align*}
    P_{\mathcal{M}}(\belief \mid \mathrm{\texttt{t}}_{\belief}(\context, \langle \trig \rangle)) &>
    P_{\mathcal{M}}(\belief \mid \mathrm{\texttt{t}}_{\belief}(\context, \langle \trig, \cancel \rangle)).
\end{align*}

\paragraph{Belief update}

Lastly, we define a model's \textit{belief update accuracy} as the proportion of items for which both
\begin{align*}
    P_{\mathcal{M}}(\belief \mid \mathrm{\texttt{t}}_{\belief}(\context, \langle \trig \rangle)) 
    &> P_{\mathcal{M}}(\neg\belief \mid \mathrm{\texttt{t}}_{\belief}(\context, \langle \trig \rangle)) \quad\text{and} \\
    P_{\mathcal{M}}(\belief \mid \mathrm{\texttt{t}}_{\belief}(\context, \langle \trig \rangle)) 
    &> P_{\mathcal{M}}(\belief \mid \mathrm{\texttt{t}}_{\belief}(\context, \langle \trig, \cancel \rangle))
\end{align*}
hold, i.e., $\mathcal{M}$ both recognizes the implicature triggered by $\trig$ and its subsequent cancellation by $\cancel$.

We take these accuracy definitions to be necessary (but not necessarily sufficient) conditions of an understanding of implicature, implicature cancellation, and of the belief updates these phenomena, taken together, entail.

\paragraph{Models}

We conduct our experiments on both open and closed instruction-tuned LLMs. The open-weight LLMs we evaluate include Gemma 3 \citep[4B, 12B, 27B;][]{gemmateam2025gemma3technicalreport}, Llama 3.1 (8B, 70B), Llama 3.2 (3B), Llama 3.3 \citep[70B;][]{grattafiori2024llama3herdmodels}, Qwen 2.5 \citep[3B, 7B, 14B, 32B, 72B;][]{qwen2025qwen25technicalreport} and Qwen 3 \citep[0.6B, 1.7B, 4B, 8B, 14B, 32B;][]{yang2025qwen3technicalreport}. For Qwen 3, we also experiment with the model's reasoning functionality. The closed-source LLMs we evaluate include GPT-5.2 and GPT-5.4 with and without their reasoning functionality.\footnote{All open-weight models are downloaded from HuggingFace and all closed-source models are accessed via OpenAI's API. See \Cref{tab:model-names} for full model names.}

\paragraph{Computing $P_{\mathcal{M}}(\belief \mid \mathrm{\texttt{t}}_{\belief}(\context,
\utthistory))$}

For open-weight LLMs, $P_{\mathcal{M}}(\belief \mid \mathrm{\texttt{t}}_{\belief}(\context,
\utthistory))$ is computed by extracting the logits corresponding to each option and
renormalizing them, averaging over both order shuffles.
For closed-source models, $P_{\mathcal{M}}(\belief \mid \mathrm{\texttt{t}}_{\belief}(\context,
\utthistory))$ is computed by sampling and using the resulting relative frequencies,
sampling five times for each order shuffle.
In our experimental setup, $\utthistory$ is instantiated as a specific utterance sequence,
e.g., $\langle \trig \rangle$ to estimate $\cg(\belief \mid \context, \langle \trig
\rangle)$, or $\langle \trig, \cancel \rangle$ to estimate $\cg(\belief \mid \context,
\langle \trig, \cancel \rangle)$, allowing us to evaluate how each additional utterance
affects the common ground probability assigned to $\belief$.

\section{Implicature Recognition Experiment}
\label{sec:impli_experiments}

Here, we present the experiments and results for the implicature recognition task.

\begin{figure*}[t]
    \centering
    \includegraphics[width=\linewidth]{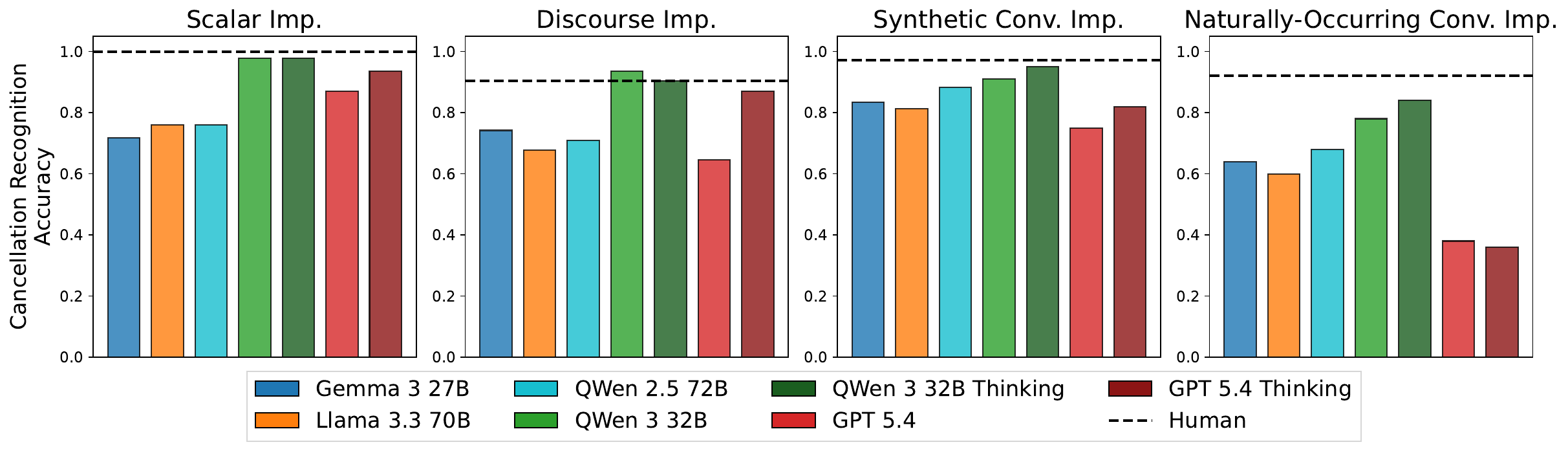}
    \caption{Cancellation recognition accuracies of the largest and most modern models for each class of models tested.
    }
    \label{fig:main:cancel_recognition}
\end{figure*}

\begin{figure*}[t]
    \centering
    \includegraphics[width=\linewidth]{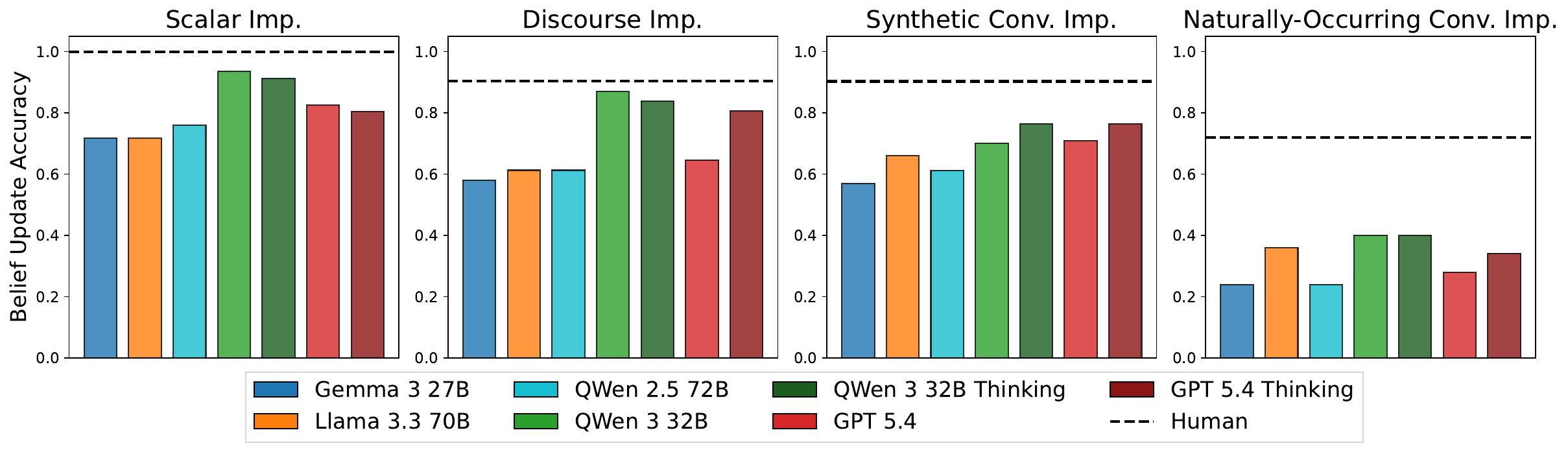}
    \caption{Belief update accuracies of the largest and most modern models for each class of models tested.
    }
    \label{fig:main:belief_update}
\end{figure*}

\subsection{Human Topline and Prior Common Ground Control}

We use our crowdsourcing annotation results from Section~\ref{sec:main:crowdsourcing_annotation} to compute an implicature recognition \textit{human accuracy topline}. To do so, we compute the proportion of items for which the z-scored Likert ratings averaged across participants are above $0$.

We run an additional experiment estimating the \textit{prior common ground} $\cgprior$ for every
item, i.e., $P_{\mathcal{M}}(\belief \mid \mathrm{\texttt{t}}_{\belief}(\emptyset, \emptyset))$,
by removing $\context$ and $\utthistory$ from the prompt and rewording the question to ask the model about its prior knowledge. 
We run this experiment to isolate the effect of the LLM's prior belief on implicature recognition. For instance, in the case of scalar
implicatures, this prior control may reveal to what extent the ``not all'' pragmatic
interpretation of ``some'' has been ``memorized''.

\subsection{Results}

The results for the implicature recognition task experiment as well as the prior belief control experiment are shown in Figure~\ref{fig:main:impli_prior_contol}.
Overall, we see that models are able to recognize scalar implicatures and discourse implicatures similarly to humans with Gemma 3 27B matching human scalar implicature recognition accuracy of $1.0$ and Llama 3.3 70B outperforming the human discourse implicature recognition accuracy of $0.90$ by $0.04$. 
On the other hand, models tend to perform worse on conversational implicatures. 
Compared to the human accuracies of $0.91$ and $0.78$ on synthetic and naturally-occurring implicatures, the best-performing models achieve implicature recognition accuracies of $0.81$ (GPT-5.4 Thinking) and $0.60$ (Llama 3.3 70B), respectively. 
Furthermore, all of the models except for Llama 3.3 70B perform worse than random at identifying naturally-occurring implicatures, demonstrating that understanding unspoken beliefs remains a challenging task for LLMs. Full results are shown in Table~\ref{tab:implicature_acc} (Appendix~\ref{sec:app:results}).

\paragraph{LLMs have strong priors for certain types of implicatures}

In addition, our prior belief experiment reveals that models have a moderately strong prior for synthetic conversational implicatures and an extremely strong prior for \someall scalar implicatures. For instance, both Llama 3.3 70B and Qwen 2.5 72B are able to achieve perfect implicature recognition accuracy on scalar implicature items \emph{without} having access to their corresponding utterances.
This result suggests that the performance of models on implicature recognition may not just stem from an understanding of implicated beliefs, but also from the prior knowledge these models may have about these implicated beliefs. 
We leave a thorough investigation of the interaction between LLMs' pragmatic understanding and their prior knowledge to future work.

\section{Cancellation Recognition and Belief Update Experiment}
\label{sec:cancel_experiments}

Next, we present the experiments and results for the cancellation recognition and belief update tasks.

\subsection{Human Topline and Controls}

We use our crowdsourcing annotation results from Section~\ref{sec:main:crowdsourcing_annotation} to compute cancellation recognition and belief update \textit{human accuracy topline}. 
For the cancellation accuracy, we compute the proportion of items for which the average per-item z-scored Likert rating decreases when the cancelling utterance is introduced. 
For the \textit{belief update accuracy}, we compute the proportion of items for which the average per-item z-scored Likert rating is above $0$ given the triggering utterance and subsequently decreases given the cancelling utterance.

\paragraph{Form control}

We evaluate the extent to which the form of the cancellation affects an LLM's success in
performing cancellation recognition. 
To do so, we create a new set of stimuli,
\negdataset, in which cancelling utterances have the form
$\explicit = $``$d + \neg\belief$'' where $d$ is a discourse marker (e.g., \textit{in fact}, \textit{actually}, etc.) and $\neg\belief$ denotes an explicit negation of the implicated belief $\belief$ triggered by $\trig$.

\paragraph{Update type control}

We also evaluate the extent to which LLM belief update accuracy is affected by the type of
follow-up utterance, i.e., whether the utterance that follows $\trig$ cancels, strengthens, or leaves $\belief$ unchanged.
To do so, we create two new sets of stimuli: (1)~\strengthdataset, in which
cancelling utterances $\cancel$ are replaced with strengthening utterances $\strength$,
i.e., utterances that strengthen $\belief$ by providing information reinforcing its plausibility, and (2)~\neutraldataset, in which $\cancel$ is replaced with a randomly sampled utterance $\neutral$ of the same stimuli type that leaves the belief $\belief$ unchanged. 
We evaluate a model's ability to perform belief strengthening by computing the proportion of \strengthdataset items for which $\probstrength > \probimp$, and its ability to leave its belief unchanged by computing the proportion of \neutraldataset items for which%
\begin{align*}
    \probneutral &> \probnegneutral \\
    &\iff \\
    \probimp &> \probnegimp.
\end{align*}%

We provide examples for the different experimental controls in Table~\ref{main:tab:running_example}. Additional details regarding the composition of these control datasets can be found in \Cref{sec:app:control_datasets}.

\begin{table}[h]
\centering
\footnotesize
\setlength{\tabcolsep}{4pt}
\renewcommand{\arraystretch}{1.3}
\begin{tabular}{p{2.3cm} p{4.7cm}}
\toprule
Control Type & Follow-Up Utterance $\cancelcolor{u}$ \\
\midrule
\quad Original      & $\cancel$: ``Though they must be done by now.'' \\
\quad Strengthening    & $\strength$: ``Maybe go over and ask them?'' \\
\quad Unchanging   & $\neutral$: ``Thankfully, I submitted before my laptop broke.'' \\
\quad Negation & $\explicit$: ``Though, to be clear, I'm not saying that you could help by doing $X$ for $C$.'' \\
\bottomrule
\end{tabular}
\caption{Control conditions for the belief update task, with fixed context \contextstmt{Do you need help with anything?}, utterance \trigstmt{I think $C$ was looking for someone to do $X$.} and belief \beliefstmt{You can help by doing $X$ for $C$.}}
\label{main:tab:running_example}
\end{table}

\subsection{Results}

We present model performance on cancellation recognition in Figure~\ref{fig:main:cancel_recognition}. Overall, LLMs perform cancellation recognition at levels close to human performance, with some models slightly surpassing human cancellation recognition accuracy---e.g., Qwen 3 32B achieves a cancellation accuracy of $0.94$ on discourse implicatures whereas humans achieve an accuracy of $0.90$. However, models perform substantially worse when presented with naturally-occurring cancellations. In this setting, the best model, Qwen 3 32B Thinking, achieves a cancellation recognition accuracy of $0.84$, compared to human performance of $0.92$. 

We also report results for belief updating in Figure~\ref{fig:main:belief_update}. When evaluating the joint tasks of implicature and cancellation recognition, model performance falls sharply, especially for conversational implicatures. For instance, the model with highest belief update accuracy, Qwen 3 32B Thinking, achieves belief update accuracies of $0.76$ and $0.40$ on synthetic and naturally-occurring conversational implicatures, respectively, whereas humans achieve $0.90$ and $0.72$. Full results can be found in Tables~\ref{tab:decrease_acc}~and~\ref{tab:impli_decrease_acc} in Appendix~\ref{sec:app:results}.

\begin{figure}[t]
    \centering
    \includegraphics[width=\linewidth]{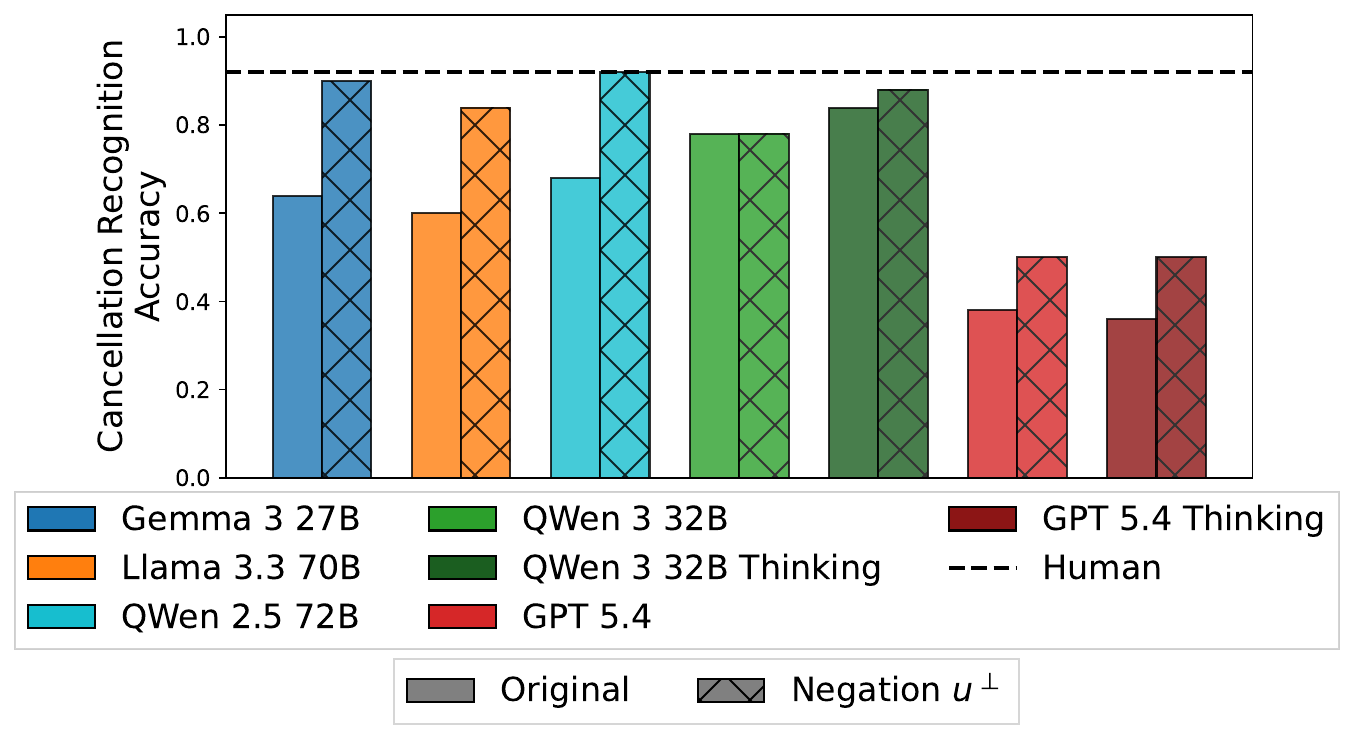}
    \caption{Cancellation recognition accuracy of models on the naturally-occurring conversational implicatures using the original cancelling utterances ($\cancel$) and the cancelling utterances with explicit negation ($\explicit$).}
    \label{fig:main:naturally_occurring_w_negation}
\end{figure}

\begin{figure}[t]
    \centering
    \includegraphics[width=\linewidth]{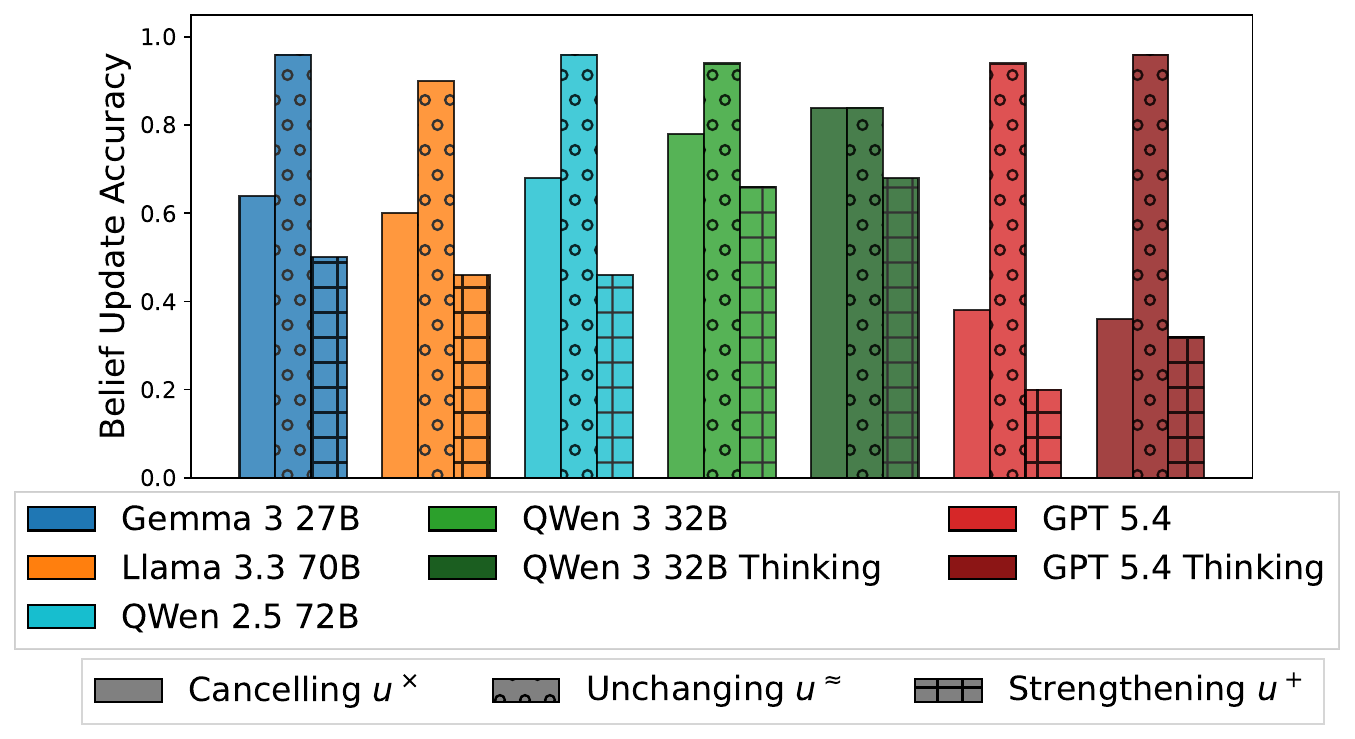}
    \caption{Update recognition accuracy of models on the naturally-occurring conversational implicatures using follow-up utterances of different types: cancelling ($\cancel$), unchanging ($\neutral$), and strengthening ($\strength$).}
    \label{fig:main:naturally_occurring_w_types}
\end{figure}

\paragraph{Explicit negation does not lead to perfect LLM cancellation recognition}

We compare model cancellation recognition accuracy when using the original \datasetname stimuli and the \negdataset variant, which contain explicitly negating cancelling utterances. We show the cancellation recognition accuracies on the naturally-occurring conversational implicatures in Figure~\ref{fig:main:naturally_occurring_w_negation} and on the full set of implicature types in Figure~\ref{fig:app:belief_update_infact} in Appendix~\ref{sec:app:results}. While the implicature cancellations with explicit negations lead to higher cancellation recognition accuracies, we find that in most cases --- aside from scalar implicatures --- the cancellation recognition accuracy falls short of its theoretical upper bound of one. This suggests that factors beyond pragmatic competence, such as difficulties in negation interpretation, may also contribute to observed cancellation recognition errors.

\paragraph{LLMs are better at maintaining existing beliefs than changing them}

We show the belief update accuracies using the cancelling, strengthening and unchanging utterances on naturally-occurring conversational implicatures in Figure~\ref{fig:main:naturally_occurring_w_types} and on the entire set of implicature types in Figure~\ref{fig:app:belief_update_by_type} in Appendix~\ref{sec:app:results}. Overall, we observe that LLMs struggle to update their beliefs both with strengthening and cancelling utterances. However, they have relatively little difficulty maintaining their beliefs when presented with an irrelevant follow-up utterance. For instance, while GPT-5.4 maintains its belief with an accuracy of $0.96$ on naturally-occurring conversational implicatures, its update accuracies when presented with strengthening and cancelling utterances are both at $0.32$. This result suggests that current LLMs are better at maintaining beliefs than at changing them.

\section{Conclusion}
\label{sec:conclusion}

In conclusion, in this paper we study LLM understanding of communicative beliefs and belief updates via implicature and cancellation recognition. To this end, we develop the first implicature cancellation dataset, \datasetname, and show that LLMs fall short of a human understanding of unspoken beliefs and belief updates. Control experiments reveal key weaknesses; for instance, a reliance on prior biases, an inability to reconcile explicit updates and a dependence on update type. Our results highlight a critical gap between current LLM capabilities and human-level pragmatic reasoning which calls for further research into how models manage evolving communicative beliefs.

\section*{Limitations}

We identify three main limitations with our work. Firstly, although we thoroughly reviewed whether the implicatures and corresponding cancellations are, in fact, annotated as such by both experts and lay annotators, whether or not an utterance cancels a prior statement remains open to interpretation. With five annotations per stimulus, we consider our results to be reliable, but we encourage future work to further explore this, particularly focusing on the examples for which our annotators demonstrate disagreement. Expanding the number of annotations could clarify whether or not such disagreement is due to legitimate stimulus ambiguity or instead to noise in our data.

Secondly, we identified that models' implicature recognition performance is strongly affected by their prior beliefs: without knowing utterance $\trig$\ for $\langle$\textit{some}, \textit{all}$\rangle$ statements, they will promptly score corresponding belief $\belief$\ highly. As a result, we cannot say with certainty that their implicature recognition accuracy is not affected by this; we leave disentangling scalar implicature reasoning from prior beliefs to future work.

Lastly, we elicited LLM judgements by extracting probabilities from their output distributions. Alternatively, one could provide the LLM with utterances and cancelling utterances, and ask them to further expand the provided discourse. Consistency vs. inconsistency with the beliefs implicitly conveyed in such generated text could reveal whether the LLMs \textit{really} manage to remain consistent with those beliefs. We did not opt for this type of evaluation due to its lack of scalability, i.e., high-quality human annotations of generated text would be needed to assess the nuances behind the LLM responses. 

\section*{Acknowledgements}

The authors would like to thank the reviewers for their valuable comments. We would also like to thank Gaurav Kamath and Austin Kraft for their helpful comments on an earlier version of this paper. This work was supported by the Fonds de Recherche du Québec – Nature et Technologies (FRQNT), the Natural Sciences and Engineering Research Council of Canada (NSERC), the IVADO Postdoctoral Research Funding Program, and the Canada CIFAR AI Chair Program. We acknowledge material support from NVIDIA Corporation in the form of computational resources provided to Mila.

\bibliography{custom}

@misc{yang2025qwen3technicalreport,
      title={{Qwen3} Technical Report}, 
      author={An Yang and Anfeng Li and Baosong Yang and Beichen Zhang and Binyuan Hui and Bo Zheng and Bowen Yu and Chang Gao and Chengen Huang and Chenxu Lv and Chujie Zheng and Dayiheng Liu and Fan Zhou and Fei Huang and Feng Hu and Hao Ge and Haoran Wei and Huan Lin and Jialong Tang and Jian Yang and Jianhong Tu and Jianwei Zhang and Jianxin Yang and Jiaxi Yang and Jing Zhou and Jingren Zhou and Junyang Lin and Kai Dang and Keqin Bao and Kexin Yang and Le Yu and Lianghao Deng and Mei Li and Mingfeng Xue and Mingze Li and Pei Zhang and Peng Wang and Qin Zhu and Rui Men and Ruize Gao and Shixuan Liu and Shuang Luo and Tianhao Li and Tianyi Tang and Wenbiao Yin and Xingzhang Ren and Xinyu Wang and Xinyu Zhang and Xuancheng Ren and Yang Fan and Yang Su and Yichang Zhang and Yinger Zhang and Yu Wan and Yuqiong Liu and Zekun Wang and Zeyu Cui and Zhenru Zhang and Zhipeng Zhou and Zihan Qiu},
      year={2025},
      eprint={2505.09388},
      archivePrefix={arXiv},
      primaryClass={cs.CL},
      url={https://arxiv.org/abs/2505.09388}, 
}

@misc{qwen2025qwen25technicalreport,
      title={{Qwen2.5} Technical Report}, 
      author={Qwen-Team and An Yang and Baosong Yang and Beichen Zhang and Binyuan Hui and Bo Zheng and Bowen Yu and Chengyuan Li and Dayiheng Liu and Fei Huang and Haoran Wei and Huan Lin and Jian Yang and Jianhong Tu and Jianwei Zhang and Jianxin Yang and Jiaxi Yang and Jingren Zhou and Junyang Lin and Kai Dang and Keming Lu and Keqin Bao and Kexin Yang and Le Yu and Mei Li and Mingfeng Xue and Pei Zhang and Qin Zhu and Rui Men and Runji Lin and Tianhao Li and Tianyi Tang and Tingyu Xia and Xingzhang Ren and Xuancheng Ren and Yang Fan and Yang Su and Yichang Zhang and Yu Wan and Yuqiong Liu and Zeyu Cui and Zhenru Zhang and Zihan Qiu},
      year={2025},
      eprint={2412.15115},
      archivePrefix={arXiv},
      primaryClass={cs.CL},
      url={https://arxiv.org/abs/2412.15115}, 
}

@misc{grattafiori2024llama3herdmodels,
      title={The {Llama} 3 Herd of Models}, 
      author={Aaron Grattafiori and Abhimanyu Dubey and Abhinav Jauhri and Abhinav Pandey and Abhishek Kadian and Ahmad Al-Dahle and Aiesha Letman and Akhil Mathur and Alan Schelten and Alex Vaughan and Amy Yang and Angela Fan and Anirudh Goyal and Anthony Hartshorn and Aobo Yang and Archi Mitra and Archie Sravankumar and Artem Korenev and Arthur Hinsvark and Arun Rao and Aston Zhang and Aurelien Rodriguez and Austen Gregerson and Ava Spataru and Baptiste Roziere and Bethany Biron and Binh Tang and Bobbie Chern and Charlotte Caucheteux and Chaya Nayak and Chloe Bi and Chris Marra and Chris McConnell and Christian Keller and Christophe Touret and Chunyang Wu and Corinne Wong and Cristian Canton Ferrer and Cyrus Nikolaidis and Damien Allonsius and Daniel Song and Danielle Pintz and Danny Livshits and Danny Wyatt and David Esiobu and Dhruv Choudhary and Dhruv Mahajan and Diego Garcia-Olano and Diego Perino and Dieuwke Hupkes and Egor Lakomkin and Ehab AlBadawy and Elina Lobanova and Emily Dinan and Eric Michael Smith and Filip Radenovic and Francisco Guzmán and Frank Zhang and Gabriel Synnaeve and Gabrielle Lee and Georgia Lewis Anderson and Govind Thattai and Graeme Nail and Gregoire Mialon and Guan Pang and Guillem Cucurell and Hailey Nguyen and Hannah Korevaar and Hu Xu and Hugo Touvron and Iliyan Zarov and Imanol Arrieta Ibarra and Isabel Kloumann and Ishan Misra and Ivan Evtimov and Jack Zhang and Jade Copet and Jaewon Lee and Jan Geffert and Jana Vranes and Jason Park and Jay Mahadeokar and Jeet Shah and Jelmer van der Linde and Jennifer Billock and Jenny Hong and Jenya Lee and Jeremy Fu and Jianfeng Chi and Jianyu Huang and Jiawen Liu and Jie Wang and Jiecao Yu and Joanna Bitton and Joe Spisak and Jongsoo Park and Joseph Rocca and Joshua Johnstun and Joshua Saxe and Junteng Jia and Kalyan Vasuden Alwala and Karthik Prasad and Kartikeya Upasani and Kate Plawiak and Ke Li and Kenneth Heafield and Kevin Stone and Khalid El-Arini and Krithika Iyer and Kshitiz Malik and Kuenley Chiu and Kunal Bhalla and Kushal Lakhotia and Lauren Rantala-Yeary and Laurens van der Maaten and Lawrence Chen and Liang Tan and Liz Jenkins and Louis Martin and Lovish Madaan and Lubo Malo and Lukas Blecher and Lukas Landzaat and Luke de Oliveira and Madeline Muzzi and Mahesh Pasupuleti and Mannat Singh and Manohar Paluri and Marcin Kardas and Maria Tsimpoukelli and Mathew Oldham and Mathieu Rita and Maya Pavlova and Melanie Kambadur and Mike Lewis and Min Si and Mitesh Kumar Singh and Mona Hassan and Naman Goyal and Narjes Torabi and Nikolay Bashlykov and Nikolay Bogoychev and Niladri Chatterji and Ning Zhang and Olivier Duchenne and Onur Çelebi and Patrick Alrassy and Pengchuan Zhang and Pengwei Li and Petar Vasic and Peter Weng and Prajjwal Bhargava and Pratik Dubal and Praveen Krishnan and Punit Singh Koura and Puxin Xu and Qing He and Qingxiao Dong and Ragavan Srinivasan and Raj Ganapathy and Ramon Calderer and Ricardo Silveira Cabral and Robert Stojnic and Roberta Raileanu and Rohan Maheswari and Rohit Girdhar and Rohit Patel and Romain Sauvestre and Ronnie Polidoro and Roshan Sumbaly and Ross Taylor and Ruan Silva and Rui Hou and Rui Wang and Saghar Hosseini and Sahana Chennabasappa and Sanjay Singh and Sean Bell and Seohyun Sonia Kim and Sergey Edunov and Shaoliang Nie and Sharan Narang and Sharath Raparthy and Sheng Shen and Shengye Wan and Shruti Bhosale and Shun Zhang and Simon Vandenhende and Soumya Batra and Spencer Whitman and Sten Sootla and Stephane Collot and Suchin Gururangan and Sydney Borodinsky and Tamar Herman and Tara Fowler and Tarek Sheasha and Thomas Georgiou and Thomas Scialom and Tobias Speckbacher and Todor Mihaylov and Tong Xiao and Ujjwal Karn and Vedanuj Goswami and Vibhor Gupta and Vignesh Ramanathan and Viktor Kerkez and Vincent Gonguet and Virginie Do and Vish Vogeti and Vítor Albiero and Vladan Petrovic and Weiwei Chu and Wenhan Xiong and Wenyin Fu and Whitney Meers and Xavier Martinet and Xiaodong Wang and Xiaofang Wang and Xiaoqing Ellen Tan and Xide Xia and Xinfeng Xie and Xuchao Jia and Xuewei Wang and Yaelle Goldschlag and Yashesh Gaur and Yasmine Babaei and Yi Wen and Yiwen Song and Yuchen Zhang and Yue Li and Yuning Mao and Zacharie Delpierre Coudert and Zheng Yan and Zhengxing Chen and Zoe Papakipos and Aaditya Singh and Aayushi Srivastava and Abha Jain and Adam Kelsey and Adam Shajnfeld and Adithya Gangidi and Adolfo Victoria and Ahuva Goldstand and Ajay Menon and Ajay Sharma and Alex Boesenberg and Alexei Baevski and Allie Feinstein and Amanda Kallet and Amit Sangani and Amos Teo and Anam Yunus and Andrei Lupu and Andres Alvarado and Andrew Caples and Andrew Gu and Andrew Ho and Andrew Poulton and Andrew Ryan and Ankit Ramchandani and Annie Dong and Annie Franco and Anuj Goyal and Aparajita Saraf and Arkabandhu Chowdhury and Ashley Gabriel and Ashwin Bharambe and Assaf Eisenman and Azadeh Yazdan and Beau James and Ben Maurer and Benjamin Leonhardi and Bernie Huang and Beth Loyd and Beto De Paola and Bhargavi Paranjape and Bing Liu and Bo Wu and Boyu Ni and Braden Hancock and Bram Wasti and Brandon Spence and Brani Stojkovic and Brian Gamido and Britt Montalvo and Carl Parker and Carly Burton and Catalina Mejia and Ce Liu and Changhan Wang and Changkyu Kim and Chao Zhou and Chester Hu and Ching-Hsiang Chu and Chris Cai and Chris Tindal and Christoph Feichtenhofer and Cynthia Gao and Damon Civin and Dana Beaty and Daniel Kreymer and Daniel Li and David Adkins and David Xu and Davide Testuggine and Delia David and Devi Parikh and Diana Liskovich and Didem Foss and Dingkang Wang and Duc Le and Dustin Holland and Edward Dowling and Eissa Jamil and Elaine Montgomery and Eleonora Presani and Emily Hahn and Emily Wood and Eric-Tuan Le and Erik Brinkman and Esteban Arcaute and Evan Dunbar and Evan Smothers and Fei Sun and Felix Kreuk and Feng Tian and Filippos Kokkinos and Firat Ozgenel and Francesco Caggioni and Frank Kanayet and Frank Seide and Gabriela Medina Florez and Gabriella Schwarz and Gada Badeer and Georgia Swee and Gil Halpern and Grant Herman and Grigory Sizov and Guangyi and Zhang and Guna Lakshminarayanan and Hakan Inan and Hamid Shojanazeri and Han Zou and Hannah Wang and Hanwen Zha and Haroun Habeeb and Harrison Rudolph and Helen Suk and Henry Aspegren and Hunter Goldman and Hongyuan Zhan and Ibrahim Damlaj and Igor Molybog and Igor Tufanov and Ilias Leontiadis and Irina-Elena Veliche and Itai Gat and Jake Weissman and James Geboski and James Kohli and Janice Lam and Japhet Asher and Jean-Baptiste Gaya and Jeff Marcus and Jeff Tang and Jennifer Chan and Jenny Zhen and Jeremy Reizenstein and Jeremy Teboul and Jessica Zhong and Jian Jin and Jingyi Yang and Joe Cummings and Jon Carvill and Jon Shepard and Jonathan McPhie and Jonathan Torres and Josh Ginsburg and Junjie Wang and Kai Wu and Kam Hou U and Karan Saxena and Kartikay Khandelwal and Katayoun Zand and Kathy Matosich and Kaushik Veeraraghavan and Kelly Michelena and Keqian Li and Kiran Jagadeesh and Kun Huang and Kunal Chawla and Kyle Huang and Lailin Chen and Lakshya Garg and Lavender A and Leandro Silva and Lee Bell and Lei Zhang and Liangpeng Guo and Licheng Yu and Liron Moshkovich and Luca Wehrstedt and Madian Khabsa and Manav Avalani and Manish Bhatt and Martynas Mankus and Matan Hasson and Matthew Lennie and Matthias Reso and Maxim Groshev and Maxim Naumov and Maya Lathi and Meghan Keneally and Miao Liu and Michael L. Seltzer and Michal Valko and Michelle Restrepo and Mihir Patel and Mik Vyatskov and Mikayel Samvelyan and Mike Clark and Mike Macey and Mike Wang and Miquel Jubert Hermoso and Mo Metanat and Mohammad Rastegari and Munish Bansal and Nandhini Santhanam and Natascha Parks and Natasha White and Navyata Bawa and Nayan Singhal and Nick Egebo and Nicolas Usunier and Nikhil Mehta and Nikolay Pavlovich Laptev and Ning Dong and Norman Cheng and Oleg Chernoguz and Olivia Hart and Omkar Salpekar and Ozlem Kalinli and Parkin Kent and Parth Parekh and Paul Saab and Pavan Balaji and Pedro Rittner and Philip Bontrager and Pierre Roux and Piotr Dollar and Polina Zvyagina and Prashant Ratanchandani and Pritish Yuvraj and Qian Liang and Rachad Alao and Rachel Rodriguez and Rafi Ayub and Raghotham Murthy and Raghu Nayani and Rahul Mitra and Rangaprabhu Parthasarathy and Raymond Li and Rebekkah Hogan and Robin Battey and Rocky Wang and Russ Howes and Ruty Rinott and Sachin Mehta and Sachin Siby and Sai Jayesh Bondu and Samyak Datta and Sara Chugh and Sara Hunt and Sargun Dhillon and Sasha Sidorov and Satadru Pan and Saurabh Mahajan and Saurabh Verma and Seiji Yamamoto and Sharadh Ramaswamy and Shaun Lindsay and Shaun Lindsay and Sheng Feng and Shenghao Lin and Shengxin Cindy Zha and Shishir Patil and Shiva Shankar and Shuqiang Zhang and Shuqiang Zhang and Sinong Wang and Sneha Agarwal and Soji Sajuyigbe and Soumith Chintala and Stephanie Max and Stephen Chen and Steve Kehoe and Steve Satterfield and Sudarshan Govindaprasad and Sumit Gupta and Summer Deng and Sungmin Cho and Sunny Virk and Suraj Subramanian and Sy Choudhury and Sydney Goldman and Tal Remez and Tamar Glaser and Tamara Best and Thilo Koehler and Thomas Robinson and Tianhe Li and Tianjun Zhang and Tim Matthews and Timothy Chou and Tzook Shaked and Varun Vontimitta and Victoria Ajayi and Victoria Montanez and Vijai Mohan and Vinay Satish Kumar and Vishal Mangla and Vlad Ionescu and Vlad Poenaru and Vlad Tiberiu Mihailescu and Vladimir Ivanov and Wei Li and Wenchen Wang and Wenwen Jiang and Wes Bouaziz and Will Constable and Xiaocheng Tang and Xiaojian Wu and Xiaolan Wang and Xilun Wu and Xinbo Gao and Yaniv Kleinman and Yanjun Chen and Ye Hu and Ye Jia and Ye Qi and Yenda Li and Yilin Zhang and Ying Zhang and Yossi Adi and Youngjin Nam and Yu and Wang and Yu Zhao and Yuchen Hao and Yundi Qian and Yunlu Li and Yuzi He and Zach Rait and Zachary DeVito and Zef Rosnbrick and Zhaoduo Wen and Zhenyu Yang and Zhiwei Zhao and Zhiyu Ma},
      year={2024},
      eprint={2407.21783},
      archivePrefix={arXiv},
      primaryClass={cs.AI},
      url={https://arxiv.org/abs/2407.21783}, 
}

@misc{gemmateam2025gemma3technicalreport,
      title={{Gemma} 3 Technical Report}, 
      author={Gemma-Team and Aishwarya Kamath and Johan Ferret and Shreya Pathak and Nino Vieillard and Ramona Merhej and Sarah Perrin and Tatiana Matejovicova and Alexandre Ramé and Morgane Rivière and Louis Rouillard and Thomas Mesnard and Geoffrey Cideron and Jean-bastien Grill and Sabela Ramos and Edouard Yvinec and Michelle Casbon and Etienne Pot and Ivo Penchev and Gaël Liu and Francesco Visin and Kathleen Kenealy and Lucas Beyer and Xiaohai Zhai and Anton Tsitsulin and Robert Busa-Fekete and Alex Feng and Noveen Sachdeva and Benjamin Coleman and Yi Gao and Basil Mustafa and Iain Barr and Emilio Parisotto and David Tian and Matan Eyal and Colin Cherry and Jan-Thorsten Peter and Danila Sinopalnikov and Surya Bhupatiraju and Rishabh Agarwal and Mehran Kazemi and Dan Malkin and Ravin Kumar and David Vilar and Idan Brusilovsky and Jiaming Luo and Andreas Steiner and Abe Friesen and Abhanshu Sharma and Abheesht Sharma and Adi Mayrav Gilady and Adrian Goedeckemeyer and Alaa Saade and Alex Feng and Alexander Kolesnikov and Alexei Bendebury and Alvin Abdagic and Amit Vadi and András György and André Susano Pinto and Anil Das and Ankur Bapna and Antoine Miech and Antoine Yang and Antonia Paterson and Ashish Shenoy and Ayan Chakrabarti and Bilal Piot and Bo Wu and Bobak Shahriari and Bryce Petrini and Charlie Chen and Charline Le Lan and Christopher A. Choquette-Choo and CJ Carey and Cormac Brick and Daniel Deutsch and Danielle Eisenbud and Dee Cattle and Derek Cheng and Dimitris Paparas and Divyashree Shivakumar Sreepathihalli and Doug Reid and Dustin Tran and Dustin Zelle and Eric Noland and Erwin Huizenga and Eugene Kharitonov and Frederick Liu and Gagik Amirkhanyan and Glenn Cameron and Hadi Hashemi and Hanna Klimczak-Plucińska and Harman Singh and Harsh Mehta and Harshal Tushar Lehri and Hussein Hazimeh and Ian Ballantyne and Idan Szpektor and Ivan Nardini and Jean Pouget-Abadie and Jetha Chan and Joe Stanton and John Wieting and Jonathan Lai and Jordi Orbay and Joseph Fernandez and Josh Newlan and Ju-yeong Ji and Jyotinder Singh and Kat Black and Kathy Yu and Kevin Hui and Kiran Vodrahalli and Klaus Greff and Linhai Qiu and Marcella Valentine and Marina Coelho and Marvin Ritter and Matt Hoffman and Matthew Watson and Mayank Chaturvedi and Michael Moynihan and Min Ma and Nabila Babar and Natasha Noy and Nathan Byrd and Nick Roy and Nikola Momchev and Nilay Chauhan and Noveen Sachdeva and Oskar Bunyan and Pankil Botarda and Paul Caron and Paul Kishan Rubenstein and Phil Culliton and Philipp Schmid and Pier Giuseppe Sessa and Pingmei Xu and Piotr Stanczyk and Pouya Tafti and Rakesh Shivanna and Renjie Wu and Renke Pan and Reza Rokni and Rob Willoughby and Rohith Vallu and Ryan Mullins and Sammy Jerome and Sara Smoot and Sertan Girgin and Shariq Iqbal and Shashir Reddy and Shruti Sheth and Siim Põder and Sijal Bhatnagar and Sindhu Raghuram Panyam and Sivan Eiger and Susan Zhang and Tianqi Liu and Trevor Yacovone and Tyler Liechty and Uday Kalra and Utku Evci and Vedant Misra and Vincent Roseberry and Vlad Feinberg and Vlad Kolesnikov and Woohyun Han and Woosuk Kwon and Xi Chen and Yinlam Chow and Yuvein Zhu and Zichuan Wei and Zoltan Egyed and Victor Cotruta and Minh Giang and Phoebe Kirk and Anand Rao and Kat Black and Nabila Babar and Jessica Lo and Erica Moreira and Luiz Gustavo Martins and Omar Sanseviero and Lucas Gonzalez and Zach Gleicher and Tris Warkentin and Vahab Mirrokni and Evan Senter and Eli Collins and Joelle Barral and Zoubin Ghahramani and Raia Hadsell and Yossi Matias and D. Sculley and Slav Petrov and Noah Fiedel and Noam Shazeer and Oriol Vinyals and Jeff Dean and Demis Hassabis and Koray Kavukcuoglu and Clement Farabet and Elena Buchatskaya and Jean-Baptiste Alayrac and Rohan Anil and Dmitry and Lepikhin and Sebastian Borgeaud and Olivier Bachem and Armand Joulin and Alek Andreev and Cassidy Hardin and Robert Dadashi and Léonard Hussenot},
      year={2025},
      eprint={2503.19786},
      archivePrefix={arXiv},
      primaryClass={cs.CL},
      url={https://arxiv.org/abs/2503.19786}, 
}

@book{sperber1986relevance,
  author    = {Sperber, Dan and Wilson, Deirdre},
  title     = {{Relevance: {C}ommunication and Cognition}},
  year      = {1986},
}

@article{grice1957meaning,
  title={Meaning},
  author={Grice, H Paul},
  journal={The philosophical review},
  volume={66},
  number={3},
  pages={377--388},
  year={1957},
  publisher={JSTOR},
  doi={https://doi.org/10.2307/2182440},
}

@article{lewis1979scorekeeping,
  title={Scorekeeping in a language game},
  author={Lewis, David},
  journal={Journal of philosophical logic},
  volume={8},
  number={1},
  pages={339--359},
  year={1979},
  publisher={Springer},
  doi={https://doi.org/10.1007/bf00258436},
}

@article{clark1991brennan,
  title={Grounding in communication.},
  author={Clark, Herbert H and Brennan, Susan E},
  year={1991},
  publisher={American Psychological Association},
  journal = {Perspectives on Socially Shared Cognition},
  doi={https://doi.org/10.1037/10096-006},
}

@book{heim1982semantics,
  title={The semantics of definite and indefinite noun phrases},
  author={Heim, Irene Roswitha},
  year={1982},
  publisher={University of Massachusetts Amherst},
  url={https://www.proquest.com/openview/9533c80af7894f76eb795c9de2cfa18f/1?pq-origsite=gscholar&cbl=18750&diss=y}
}

@article{kamp2013theory,
  title={A theory of truth and semantic representation},
  author={Kamp, Hans},
  publisher={Formal Methods in the Study of Language},
  journal={Proceedings of the Third Amsterdam Colloquium},
  pages={277--322},
  year={1981},
  doi={https://doi.org/10.1093/oso/9780195136975.003.0013}
}

@article{stalnaker1998representation,
  title={On the representation of context},
  author={Stalnaker, Robert},
  journal={Journal of logic, language and information},
  volume={7},
  number={1},
  pages={3--19},
  year={1998},
  publisher={Springer},
  doi={https://doi.org/10.1023/a:1008254815298},
}

@phdthesis{benotti2010implicature,
  title={Implicature as an Interactive Process},
  author={Benotti, Luciana},
  year={2010},
  school={Universit{\'e} Henri Poincar{\'e}-Nancy I},
  doi={https://doi.org/10.70675/1880a7d1z6e5fz4f02z832dz3b4a0d293ebc}
}

@incollection{grice1975logic,
  title={Logic and conversation},
  author={Grice, Herbert P},
  booktitle={Speech acts},
  pages={41--58},
  year={1975},
  publisher={Brill},
  doi={https://doi.org/10.1163/9789004368811_003},
}

@article{clark1986referring,
  title={Referring as a collaborative process},
  author={Clark, Herbert H and Wilkes-Gibbs, Deanna},
  journal={Cognition},
  volume={22},
  number={1},
  pages={1--39},
  year={1986},
  publisher={Elsevier},
  doi={https://doi.org/10.1016/0010-0277(86)90010-7},
}

@article{isaacs1987references,
  title={References in conversation between experts and novices.},
  author={Isaacs, Ellen A and Clark, Herbert H},
  journal={Journal of experimental psychology: general},
  volume={116},
  number={1},
  pages={26},
  year={1987},
  publisher={American Psychological Association},
  doi={https://doi.org/10.1037/0096-3445.116.1.26},
}

@article{selten2007emergence,
  title={The emergence of simple languages in an experimental coordination game},
  author={Selten, Reinhard and Warglien, Massimo},
  journal={Proceedings of the National Academy of Sciences},
  volume={104},
  number={18},
  pages={7361--7366},
  year={2007},
  publisher={National Academy of Sciences},
  doi={https://doi.org/10.1073/pnas.0702077104},
}

@article{fetzer2018linguistic,
  author    = {Anita Fetzer},
  title     = {The linguistic realization of contrastive discourse relations in context: {C}ontextualization and discourse common ground},
  journal   = {Modélisation et utilisation du contexte},
  url       = {https://www.openscience.fr/The-Linguistic-Realization-of-Contrastive-Discourse-Relations-in-Context},
  year      = {2018},
}

@article{clark2004speaking,
  title={Speaking while monitoring addressees for understanding},
  author={Clark, Herbert H and Krych, Meredyth A},
  journal={Journal of memory and language},
  volume={50},
  number={1},
  pages={62--81},
  year={2004},
  publisher={Elsevier},
  doi={https://doi.org/10.1016/j.jml.2003.08.004}
}

@inproceedings{veinott1999video,
  title={Video helps remote work: {S}peakers who need to negotiate common ground benefit from seeing each other},
  author={Veinott, Elizabeth S and Olson, Judith and Olson, Gary M and Fu, Xiaolan},
  booktitle={Proceedings of the SIGCHI conference on Human Factors in Computing Systems},
  pages={302--309},
  year={1999},
  doi={https://doi.org/10.1145/302979.303067},
}

@article{noveck2001children,
  title={When children are more logical than adults: {E}xperimental investigations of scalar implicature},
  author={Noveck, Ira A},
  journal={Cognition},
  volume={78},
  number={2},
  pages={165--188},
  year={2001},
  publisher={Elsevier},
  doi={https://doi.org/10.1016/s0010-0277(00)00114-1}
}

@article{breheny2006generalised,
  title={{A}re generalised scalar implicatures generated by default? {A}n on-line investigation into the role of context in generating pragmatic inferences},
  author={Breheny, Richard and Katsos, Napoleon and Williams, John},
  journal={Cognition},
  volume={100},
  number={3},
  pages={434--463},
  year={2006},
  publisher={Elsevier},
  doi={https://doi.org/10.1016/j.cognition.2005.07.003},
}

@article{degen2015investigating,
  title={Investigating the distribution of some (but not all) implicatures using corpora and web-based methods},
  author={Degen, Judith},
  journal={Semantics and Pragmatics},
  volume={8},
  pages={11--1},
  year={2015},
  doi={https://doi.org/10.3765/sp.8.11},
}

@article{yang2018context,
  title={Context-sensitivity and individual differences in the derivation of scalar implicature},
  author={Yang, Xiao and Minai, Utako and Fiorentino, Robert},
  journal={Frontiers in psychology},
  volume={9},
  pages={1720},
  year={2018},
  publisher={Frontiers Media SA},
  doi={https://doi.org/10.3389/fpsyg.2018.01720},
}

@article{huang2018some,
  title={Some inferences still take time: Prosody, predictability, and the speed of scalar implicatures},
  author={Huang, Yi Ting and Snedeker, Jesse},
  journal={Cognitive psychology},
  volume={102},
  pages={105--126},
  year={2018},
  publisher={Elsevier},
  doi={https://doi.org/10.1016/j.cogpsych.2018.01.004}
}

@article{grodner2010some,
  title={“{S}ome,” and possibly all, scalar inferences are not delayed: {E}vidence for immediate pragmatic enrichment},
  author={Grodner, Daniel J and Klein, Natalie M and Carbary, Kathleen M and Tanenhaus, Michael K},
  journal={Cognition},
  volume={116},
  number={1},
  pages={42--55},
  year={2010},
  publisher={Elsevier},
  doi={https://doi.org/10.1016/j.cognition.2010.03.014},
}

@article{grosz1986attention,
  title={Attention, intentions, and the structure of discourse},
  author={Grosz, Barbara J and Sidner, Candace L},
  journal={Computational linguistics},
  volume={12},
  number={3},
  pages={175--204},
  year={1986},
  url={https://aclanthology.org/J86-3001/},
}

@article{allen1980analyzing,
  title={Analyzing intention in utterances},
  author={Allen, James F and Perrault, C Raymond},
  journal={Artificial intelligence},
  volume={15},
  number={3},
  pages={143--178},
  year={1980},
  publisher={Elsevier},
  doi={https://doi.org/10.1016/0004-3702(80)90042-9},
}

@article{dale1995computational,
  title={Computational interpretations of the {G}ricean maxims in the generation of referring expressions},
  author={Dale, Robert and Reiter, Ehud},
  journal={Cognitive science},
  volume={19},
  number={2},
  pages={233--263},
  year={1995},
  publisher={Elsevier},
  doi={https://doi.org/10.1016/0364-0213(95)90018-7},
}

@inproceedings{andreas2016reasoning,
  title={Reasoning about pragmatics with neural listeners and speakers},
  author={Andreas, Jacob and Klein, Dan},
  booktitle={Proceedings of the 2016 conference on empirical methods in natural language processing},
  pages={1173--1182},
  year={2016},
  doi={https://doi.org/10.18653/v1/d16-1125},
}

@article{korner2025common,
  title={Common ground improves learning with conversational agents},
  author={K{\"o}rner, Anita and Tolzin, Antonia and Janson, Andreas and Leimeister, Jan Marco and Rummer, Ralf},
  journal={Behaviour \& Information Technology},
  pages={1--17},
  year={2025},
  publisher={Taylor \& Francis},
  doi={https://doi.org/10.1080/0144929x.2025.2541222}
}

@inproceedings{jeretic2020natural,
  title={Are natural language inference models {IMPPRES}sive? {L}earning {IMP}licature and {PRES}upposition},
  author={Jeretic, Paloma and Warstadt, Alex and Bhooshan, Suvrat and Williams, Adina},
  booktitle={Proceedings of the 58th Annual Meeting of the Association for Computational Linguistics},
  pages={8690--8705},
  year={2020},
  doi={https://doi.org/10.18653/v1/2020.acl-main.768},
  
}

@inproceedings{cho2024pragmatic,
  title={Pragmatic inference of scalar implicature by {LLMs}},
  author={Cho, Ye-eun and mook Kim, Seong},
  booktitle={Proceedings of the 62nd Annual Meeting of the Association for Computational Linguistics (Volume 4: Student Research Workshop)},
  pages={10--20},
  year={2024},
  doi={https://doi.org/10.18653/v1/2024.acl-srw.2},
}

@article{ruis2023goldilocks,
  title={The goldilocks of pragmatic understanding: {F}ine-tuning strategy matters for implicature resolution by {LLMs}},
  author={Ruis, Laura and Khan, Akbir and Biderman, Stella and Hooker, Sara and Rockt{\"a}schel, Tim and Grefenstette, Edward},
  journal={Advances in Neural Information Processing Systems},
  volume={36},
  pages={20827--20905},
  year={2023},
  doi={https://doi.org/10.52202/075280-0913},
}

@inproceedings{yue2024large,
  title={Do Large Language Models Understand Conversational Implicature - {A} case study with a Chinese sitcom},
  author={Yue, Shisen and Song, Siyuan and Cheng, Xinyuan and Hu, Hai},
  booktitle={Proceedings of the 23rd Chinese National Conference on Computational Linguistics (Volume 1: Main Conference)},
  pages={1270--1285},
  year={2024},
  url={https://aclanthology.org/2024.ccl-1.98/}
}

@article{cong2024manner,
  title={Manner implicatures in large language models},
  author={Cong, Yan},
  journal={Scientific Reports},
  volume={14},
  number={1},
  pages={29113},
  year={2024},
  publisher={Nature Publishing Group UK London},
  doi={https://doi.org/10.1038/s41598-024-80571-3},
}

@inproceedings{lascarides1991discourse,
  title={Discourse relations and defeasible knowledge},
  author={Lascarides, Alex and Asher, Nicholas},
  booktitle={29th Annual Meeting of the Association for Computational Linguistics},
  pages={55--62},
  year={1991},
  doi={https://doi.org/10.3115/981344.981352},
}

@inproceedings{rudinger2020thinking,
  title={Thinking like a skeptic: {D}efeasible inference in natural language},
  author={Rudinger, Rachel and Shwartz, Vered and Hwang, Jena D and Bhagavatula, Chandra and Forbes, Maxwell and Le Bras, Ronan and Smith, Noah A and Choi, Yejin},
  booktitle={Findings of the Association for Computational Linguistics: EMNLP 2020},
  pages={4661--4675},
  year={2020},
  doi={https://doi.org/10.18653/v1/2020.findings-emnlp.418},
}

@inproceedings{hwang2021comet,
  title={{(Comet-)Atomic 2020:} On symbolic and neural commonsense knowledge graphs},
  author={Hwang, Jena D and Bhagavatula, Chandra and Le Bras, Ronan and Da, Jeff and Sakaguchi, Keisuke and Bosselut, Antoine and Choi, Yejin},
  booktitle={Proceedings of the AAAI conference on artificial intelligence},
  pages={6384--6392},
  year={2021},
  doi={https://doi.org/10.1609/aaai.v35i7.16792},
}

@article{deemter2012generation,
  title={Generation of referring expressions: {A}ssessing the incremental algorithm},
  author={Deemter, Kees van and Gatt, Albert and Sluis, Ielka van der and Power, Richard},
  journal={Cognitive science},
  volume={36},
  number={5},
  pages={799--836},
  year={2012},
  publisher={Wiley Online Library},
  doi={https://doi.org/10.1111/j.1551-6709.2011.01205.x},
}

@inproceedings{switchboard_corpus,
author = {Godfrey, John J. and Holliman, Edward C. and McDaniel, Jane},
title = {{SWITCHBOARD:} {T}elephone speech corpus for research and development},
year = {1992},
isbn = {0780305329},
publisher = {IEEE Computer Society},
address = {USA},
booktitle = {Proceedings of the 1992 IEEE International Conference on Acoustics, Speech and Signal Processing - Volume 1},
pages = {517–520},
numpages = {4},
location = {San Francisco, California},
series = {ICASSP'92},
doi={https://doi.org/10.1109/icassp.1992.225858}
}

@book{anderson2018essentials,
  title={Essentials of linguistics},
  author={Anderson, Catherine},
  year={2018},
  publisher={McMaster University},
  doi={https://doi.org/10.1007/978-3-476-05678-8},
}

@misc{webber2019penn,
  author    = {Prasad, Rashmi and Webber, Bonnie and Lee, Alan and Joshi, Aravind},
  title     = {{Penn Discourse Treebank Version 3.0}},
  year      = {2019},
  publisher = {Linguistic Data Consortium},
  address   = {Philadelphia},
  note      = {LDC2019T05},
  doi       = {10.35111/qebf-gk47},
  howpublished = {Web Download}
}

@article{george2020conversational,
  title={Conversational implicatures in English dialogue: {A}nnotated dataset},
  author={George, Elizabeth Jasmi and Mamidi, Radhika},
  journal={Procedia Computer Science},
  volume={171},
  pages={2316--2323},
  year={2020},
  publisher={Elsevier},
  doi={https://doi.org/10.1016/j.procs.2020.04.251},
}

@inproceedings{louis2020d,
  title={“{I}’d rather just go to bed”: {U}nderstanding Indirect Answers},
  author={Louis, Annie and Roth, Dan and Radlinski, Filip},
  booktitle={Proceedings of the 2020 Conference on Empirical Methods in Natural Language Processing (EMNLP)},
  pages={7411--7425},
  year={2020},
  doi={https://doi.org/10.18653/v1/2020.emnlp-main.601},
}

@article{wilson2021second,
  title={“{S}econd guessing yourself all the time about what they really mean...”: {C}ognitive differences between autistic and non-autistic adults in understanding implied meaning},
  author={Wilson, Alexander C and Bishop, Dorothy VM},
  journal={Autism Research},
  volume={14},
  number={1},
  pages={93--101},
  year={2021},
  publisher={Wiley Online Library},
  doi={https://doi.org/10.31234/osf.io/f6ksu},
}

@inproceedings{hu2023fine,
  title={A fine-grained comparison of pragmatic language understanding in humans and language models},
  author={Hu, Jennifer and Floyd, Sammy and Jouravlev, Olessia and Fedorenko, Evelina and Gibson, Edward},
  booktitle={Proceedings of the 61st Annual Meeting of the Association for Computational Linguistics (Volume 1: Long Papers)},
  pages={4194--4213},
  year={2023},
  doi={10.18653/v1/2023.acl-long.230}
}

@inproceedings{shi-etal-2025-judging,
    title = "{J}udging the Judges: {A} Systematic Study of Position Bias in {LLM}-as-a-Judge",
    author = "Shi, Lin  and
      Ma, Chiyu  and
      Liang, Wenhua  and
      Diao, Xingjian  and
      Ma, Weicheng  and
      Vosoughi, Soroush",
    editor = "Inui, Kentaro  and
      Sakti, Sakriani  and
      Wang, Haofen  and
      Wong, Derek F.  and
      Bhattacharyya, Pushpak  and
      Banerjee, Biplab  and
      Ekbal, Asif  and
      Chakraborty, Tanmoy  and
      Singh, Dhirendra Pratap",
    booktitle = "Proceedings of the 14th International Joint Conference on Natural Language Processing and the 4th Conference of the Asia-Pacific Chapter of the Association for Computational Linguistics",
    month = dec,
    year = "2025",
    address = "Mumbai, India",
    publisher = "The Asian Federation of Natural Language Processing and The Association for Computational Linguistics",
    doi={10.18653/v1/2025.ijcnlp-long.18},
    pages = "292--314",
    ISBN = "979-8-89176-298-5",
}

@inproceedings{kabbara-cheung-2022-investigating,
    title = "Investigating the Performance of Transformer-Based {NLI} Models on Presuppositional Inferences",
    author = "Kabbara, Jad  and
      Cheung, Jackie Chi Kit",
    editor = "Calzolari, Nicoletta  and
      Huang, Chu-Ren  and
      Kim, Hansaem  and
      Pustejovsky, James  and
      Wanner, Leo  and
      Choi, Key-Sun  and
      Ryu, Pum-Mo  and
      Chen, Hsin-Hsi  and
      Donatelli, Lucia  and
      Ji, Heng  and
      Kurohashi, Sadao  and
      Paggio, Patrizia  and
      Xue, Nianwen  and
      Kim, Seokhwan  and
      Hahm, Younggyun  and
      He, Zhong  and
      Lee, Tony Kyungil  and
      Santus, Enrico  and
      Bond, Francis  and
      Na, Seung-Hoon",
    booktitle = "Proceedings of the 29th International Conference on Computational Linguistics",
    month = oct,
    year = "2022",
    address = "Gyeongju, Republic of Korea",
    publisher = "International Committee on Computational Linguistics",
    url = "https://aclanthology.org/2022.coling-1.65/",
    pages = "779--785",

}

@article{cliff1993dominance,
  title={Dominance statistics: {O}rdinal analyses to answer ordinal questions.},
  author={Cliff, Norman},
  journal={Psychological bulletin},
  volume={114},
  number={3},
  pages={494},
  year={1993},
  publisher={American Psychological Association},
  doi={https://doi.org/10.1037/0033-2909.114.3.494},
}

@article{byrt1993bias,
  title={Bias, prevalence and kappa},
  author={Byrt, Ted and Bishop, Janet and Carlin, John B},
  journal={Journal of clinical epidemiology},
  volume={46},
  number={5},
  pages={423--429},
  year={1993},
  publisher={Elsevier},
  doi={https://doi.org/10.1016/0895-4356(93)90018-V}
}

\appendix
\crefalias{section}{appendix}
\renewcommand{\appendixname}{Appendix}
\crefname{section}{Appendix}{Appendices}

\section{Expert Annotation}
\label{sec:app:expert_annotation}

\paragraph{Additional Annotation Details}

\begin{figure}[htbp]
    \centering
    \includegraphics[width=\linewidth]{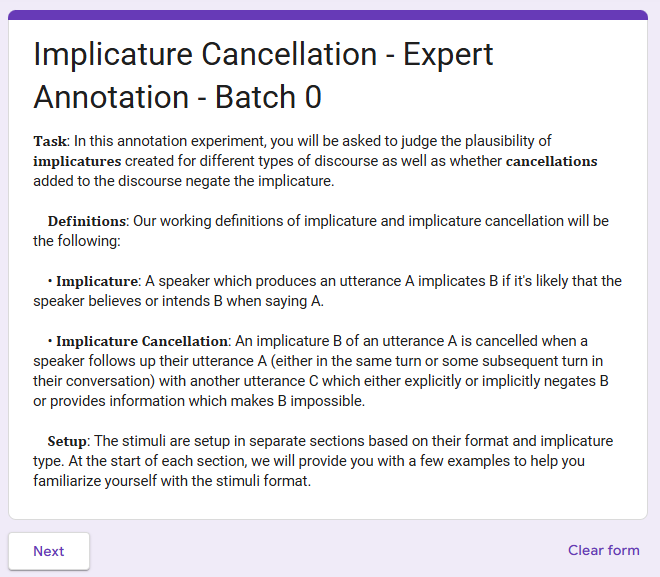}
    \caption{Landing page shown to expert annotators at the beginning of the annotation task. The landing page includes a reminder of the definitions of implicature and implicature cancellation.}
    \label{fig:front_page}
\end{figure}

\begin{figure}[htbp]
    \centering
    \includegraphics[width=\linewidth]{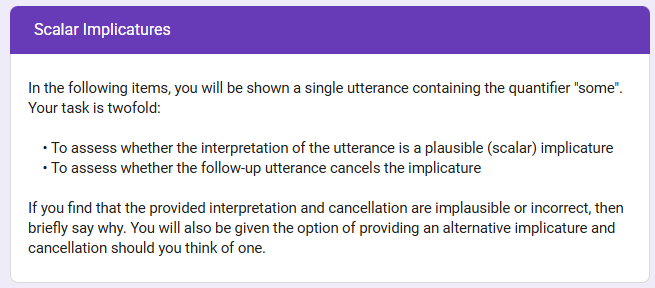}
    \caption{Instructions provided to expert annotators for the scalar implicature items.}
    \label{fig:instructions}
\end{figure}

\begin{figure}[htbp]
    \centering
    \includegraphics[width=\linewidth]{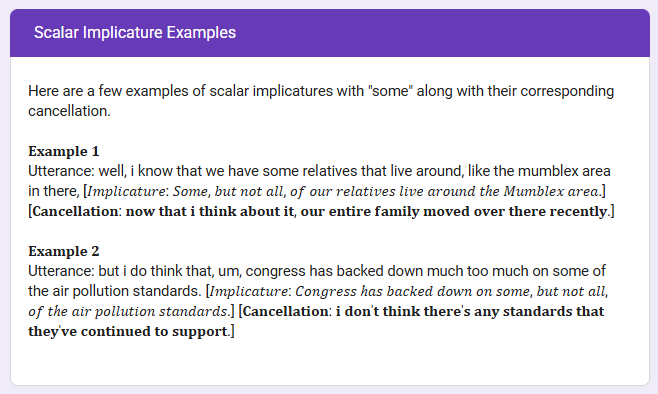}
    \caption{Example scalar implicatures shown to expert annotators prior to the annotation of the scalar implicature items. Examples like these are shown before each implicature type.}
    \label{fig:examples_provided}
\end{figure}

\begin{figure}[t]
    \centering
    \includegraphics[width=0.9\linewidth]{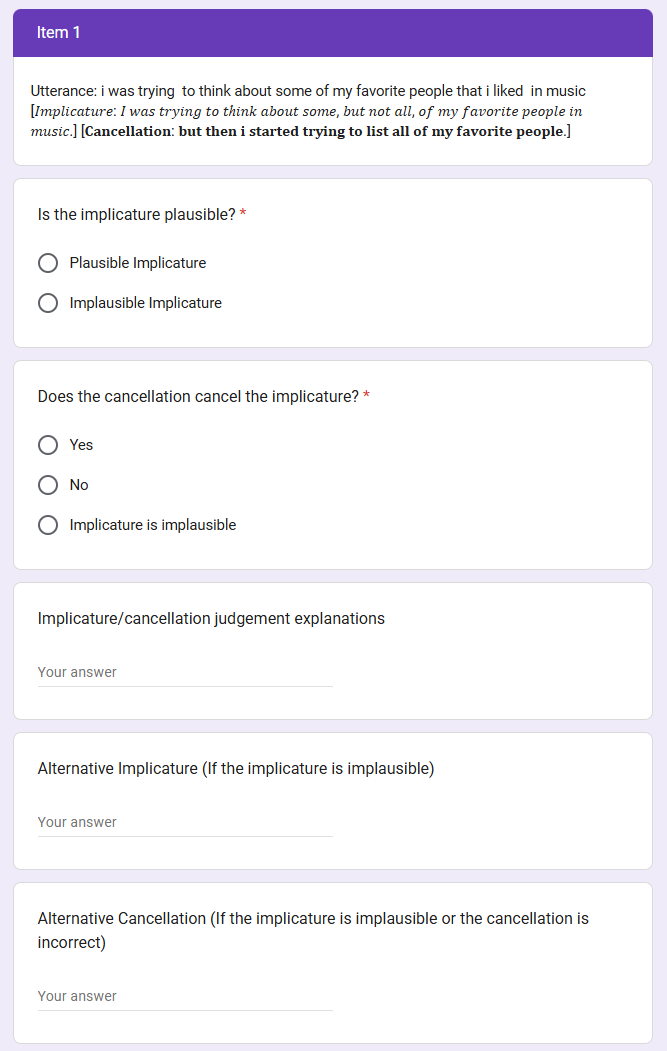}
    \caption{One of the scalar implicature items to annotate. The context, utterance, candidate implicature, and candidate implicature cancellation are shown. The expert annotator decides on the plausibility of the candidate implicature and on the correctness of the implicature cancellation.}
    \label{fig:example_item}
\end{figure}

The expert annotation interface is illustrated in Figures~\ref{fig:front_page}, \ref{fig:instructions}, \ref{fig:examples_provided}, and \ref{fig:example_item}. The expert annotators were first presented with a landing page containing general instructions along with a reminder of the notions of implicature and implicature cancellation (Figure~\ref{fig:front_page}). The annotation task was organized into several sub-batches based on stimulus type and the stimuli were presented in order of increasing cognitive load: scalar implicatures, synthetic conversational implicatures, naturally-occurring conversational implicatures, and discourse implicatures. 

Before each stimulus type, annotators are given stimulus-type-specific instructions (Figure~\ref{fig:instructions}) and representative examples to familiarize them with the stimuli and task format (Figure~\ref{fig:examples_provided}). For each annotation item (Figure~\ref{fig:example_item}), annotators are asked to judge (i) whether the proposed interpretation constitutes a plausible implicature and (ii) whether the provided cancellation is correct. If the implicature is judged implausible, annotators are asked to provide an alternative implicature. Similarly, annotators are asked to provide an alternative cancellation when the given cancellation is incorrect or when the implicature itself is deemed implausible. We elicit such alternatives for all data types except for our naturally-occurring conversational implicatures.

Expert annotators were paid at their affiliated university's teaching assistant hourly rate. This study received approval from the Research Ethics Board at McGill University (REB File \#: 21-06-019).

\paragraph{Additional Annotation Results}

We provide confusion matrices for our first round of expert annotation during which two expert annotators annotated the same batch of $53$ items. We report the two-by-two confusion matrix for the implicature plausibility responses in \Cref{tab:confusion-plausibility} and for the implicature cancellation correctness in \Cref{tab:confusion-cancellation}. For the cancellation correctness confusion matrix, we only consider the items for which there was agreement in terms of the implicature's plausibility.

\begin{table}[htbp]
    \centering
    \resizebox{0.75\columnwidth}{!}{%
    \begin{tabular}{lcc}
        \toprule
        \diagbox{$A_1$}{$A_2$}& Plausible & Implausible \\
        \midrule
        Plausible       & 43 & 4 \\
        Implausible     & 4  & 3 \\
        \bottomrule
    \end{tabular}}
    \caption{Confusion matrix of annotators $A_1$ and $A_2$ on the implicature plausibility annotation task.}
    \label{tab:confusion-plausibility}
\end{table}

\begin{table}[htbp]
    \centering
    \resizebox{0.75\columnwidth}{!}{%
    \begin{tabular}{lcc}
        \toprule
        \diagbox{$A_1$}{$A_2$} & Correct & Incorrect \\
        \midrule
        Correct         & 35 & 3 \\
        Incorrect       & 3  & 2 \\
        \bottomrule
    \end{tabular}}
    \caption{Confusion matrix of annotators $A_1$ and $A_2$ on the implicature cancellation correctness annotation task.}
    \label{tab:confusion-cancellation}
\end{table}

\section{Crowdsourcing Annotation}
\label{sec:app:crowdsourcing_annotation}

\paragraph{Additional Crowdsourcing Details}

\begin{figure}[htbp]
    \centering
    \includegraphics[width=\linewidth]{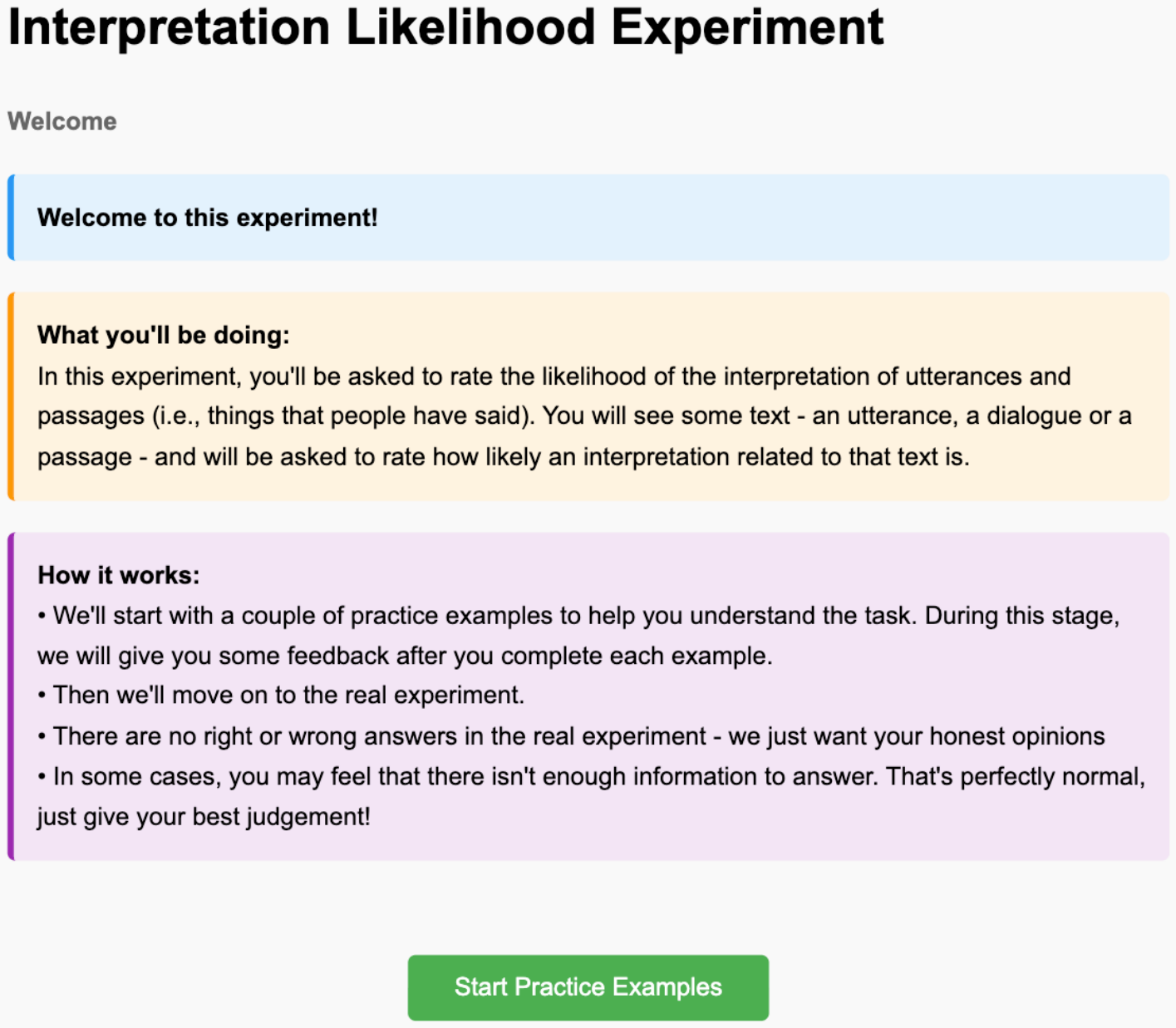}
    \caption{Landing page of the crowdsourcing experiment, outlining the instructions and interface for participants.}
    \label{fig:experiment_landing_page}
\end{figure}

\begin{figure}[htbp]
    \centering
    \includegraphics[width=\linewidth]{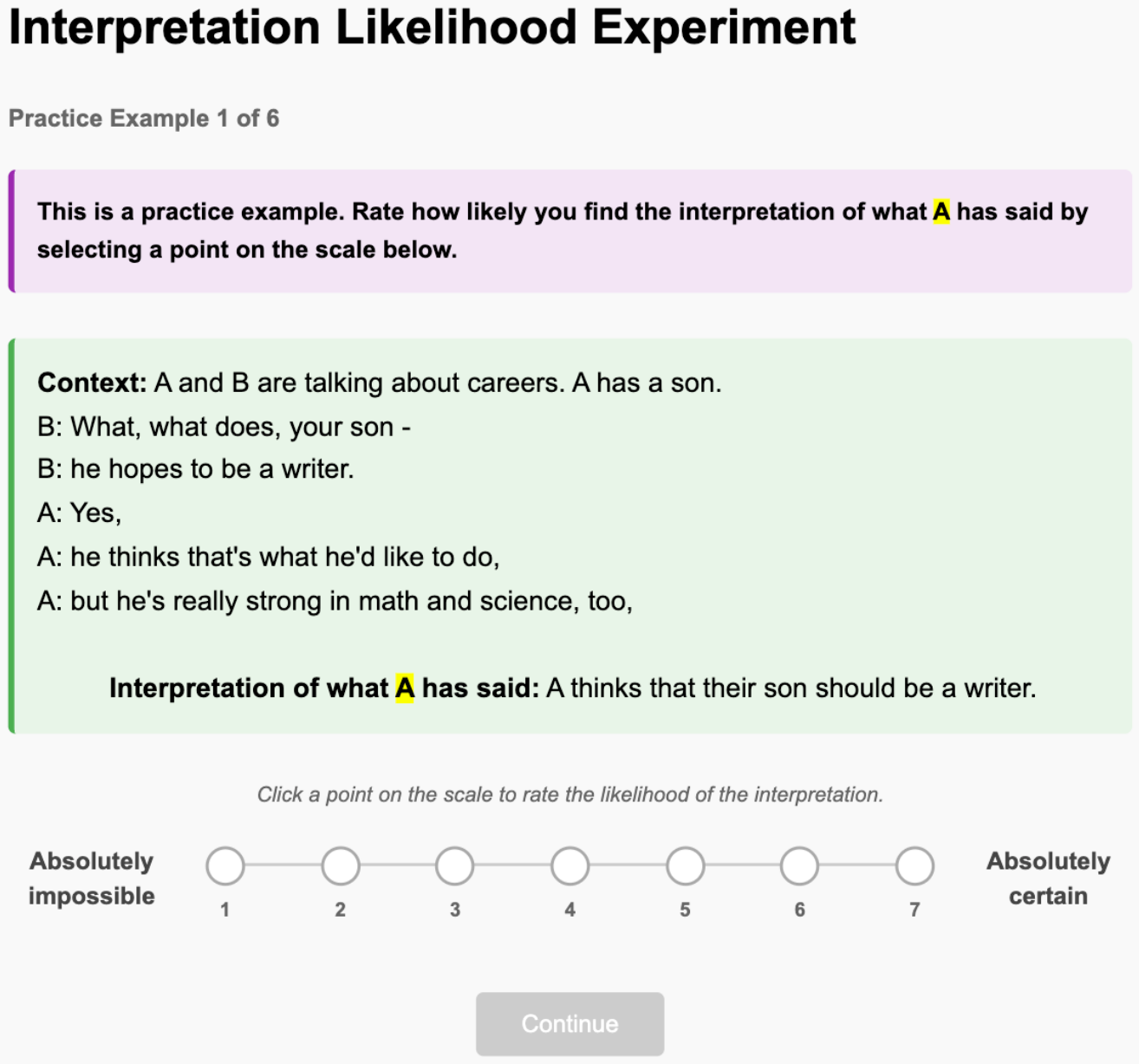}
    \caption{Example stimulus from the experiment, demonstrating the Likert-scale interface used to collect interpretation likelihood ratings.}
    \label{fig:experiment_stimulus}
\end{figure}

The crowdsourcing annotation interface is shown in Figures~\ref{fig:experiment_landing_page}~and~\ref{fig:experiment_stimulus}. Each annotator began the annotation from the same landing page (Figure~\ref{fig:experiment_landing_page}) which contains general information about the study and its structure. The crowdworkers were then shown six example stimuli (Figure~\ref{fig:experiment_stimulus}): one scalar implicature, two discourse implicatures, two synthetic conversational implicatures and one naturally-occurring conversational implicature. They were asked to try again if they provided an unlikely Likert score rating (e.g., 1, 2, or 3) for an item which did not contain a cancellation or if they provided a likely Likert score rating (e.g., 5, 6, or 7) for an item which did contain a cancellation. 

After completing the training examples, the crowdworkers were presented with either $40$ or $42$ items depending on the batch assigned to them. For the $16$ batches containing $40$ items, $10$ were attention checks (five items which should be rated likely and five items which should be rated unlikely), $15$ were items without the cancelling utterance and $15$ were items with the cancelling utterance. For the two batches of $42$ items, $10$ were attention checks, $16$ were items without the cancelling utterance and $16$ were items with the cancelling utterance. The $10$ attention checks were the same for all batches. We ensured that no batch contained the same implicature more than once.

All crowdworkers were paid 15 USD/hour via the Prolific platform\footnote{\url{https://www.prolific.com/}}. Data ingestion was done through Proliferate\footnote{\url{https://docs.proliferate.alps.science/}}. This study received approval from the Research Ethics Board at McGill University (REB File \#: 21-06-019).

\paragraph{Additional Crowdsourcing Results}

\begin{figure*}[htbp]
    \centering
    \includegraphics[width=\linewidth]{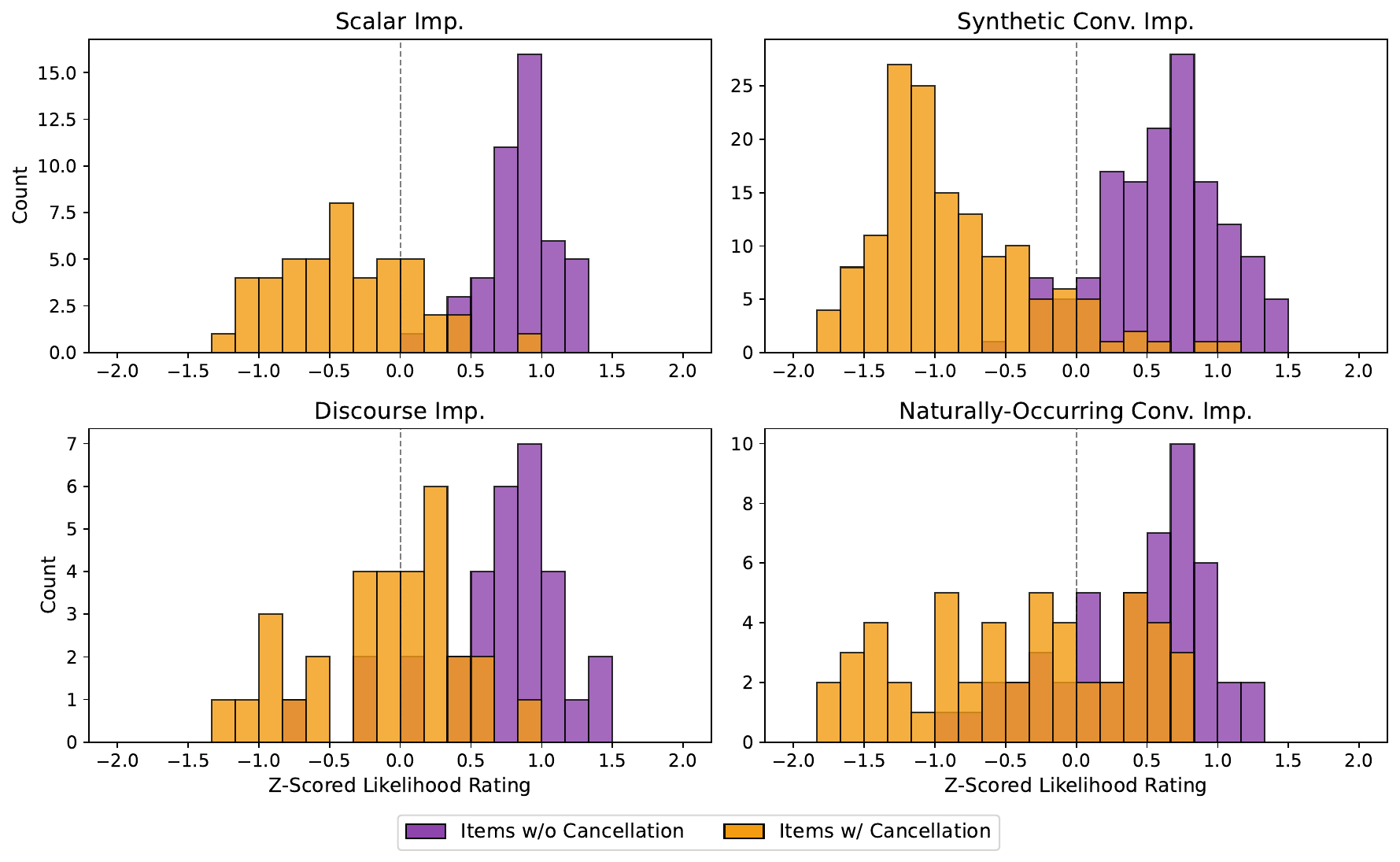}
    \caption{Disaggregated histograms of the average human likelihood z-scores with and without the cancelling utterance.}
    \label{fig:disag_human_ratings}
\end{figure*}

We provide the per-implicature-type disaggregated histograms of the average human likelihood z-scores in \Cref{fig:disag_human_ratings}. Overall, the disaggregated histograms reveal that, across all implicature types, cancelling utterances weaken the z-scored human likelihoods of implicatures.

\begin{figure*}[htbp]
    \centering
    \includegraphics[width=\linewidth]{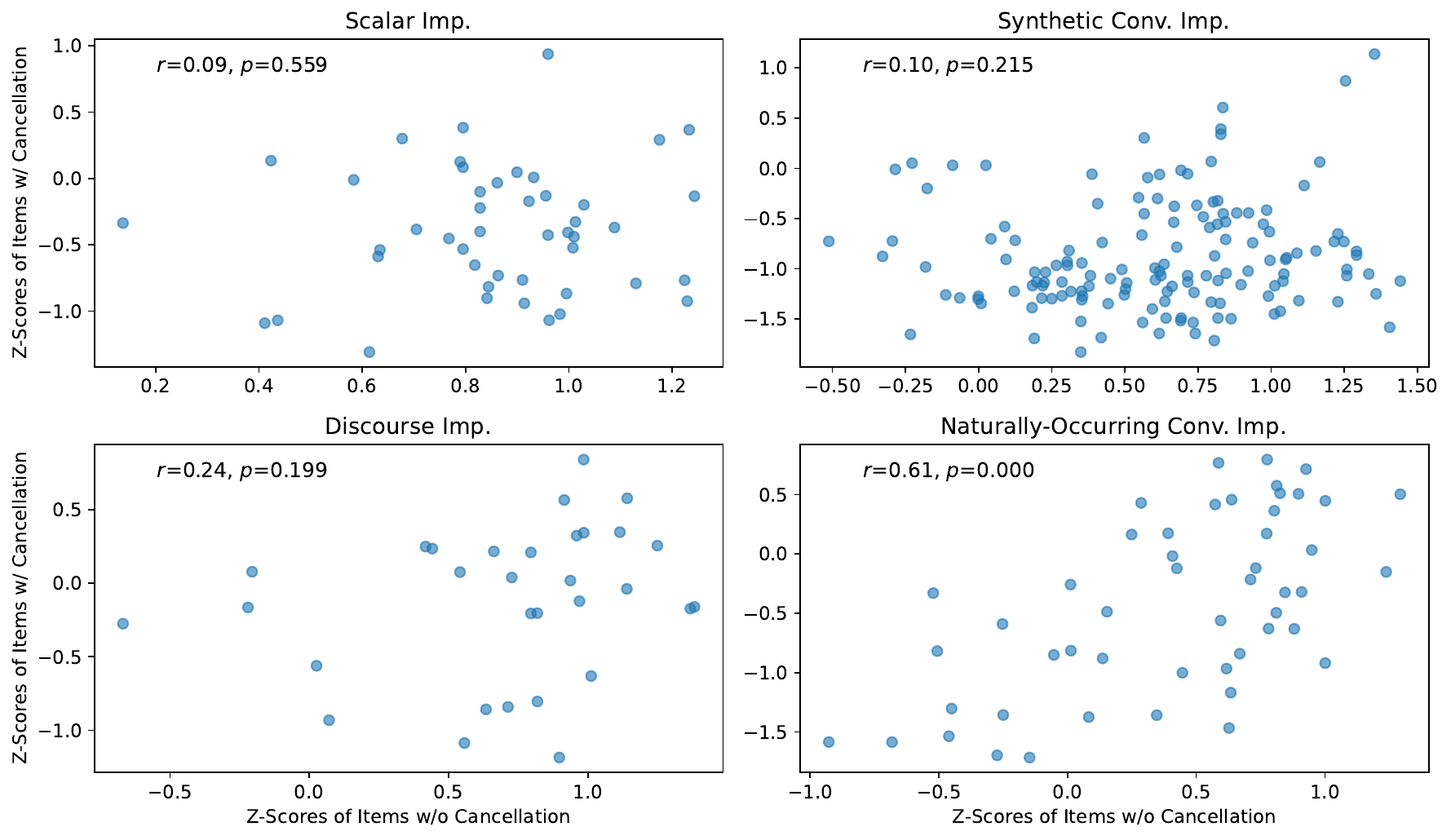}
    \caption{Scatter plots of the average human likelihood z-scores with and without the cancelling utterance. The Pearson correlation coefficient $r$ and p-value $p$ are reported at the top left.}
    \label{fig:disag_scatter_plot}
\end{figure*}

We also provide disaggregated scatter plots with the human likelihood z-scores with and without the cancelling utterance on the y-axis and x-axis respectively in \Cref{fig:disag_scatter_plot}. We also report the Pearson correlation $r$ for each of the implicature types in their corresponding subplot. Overall, we find no statistically significant correlation between an implicature's likelihood with and without a cancelling utterance for scalar implicature, discourse implicatures and synthetic conversational implicatures. However, we do find a statistically significant correlation for naturally-occurring conversational implicatures ($r=0.63$, $p<0.001$) which suggests that ``stronger'' naturally-occurring implicatures are more ``difficult'' to cancel.

\clearpage

\onecolumn

\section{Model Details}
\label{sec:app:model_details}

\begin{table*}[htbp]
\centering
\small
    \begin{tabular}{ll}
        \toprule
        \textbf{Model Name} & \textbf{HuggingFace or OpenAI Identifier} \\
        \midrule
        Gemma 3 (4B)   & \texttt{google/gemma-3-4b-it} \\
        Gemma 3 (12B)  & \texttt{google/gemma-3-12b-it} \\
        Gemma 3 (27B)  & \texttt{google/gemma-3-27b-it} \\
        \midrule
        Llama 3.1 (8B)  & \texttt{meta-llama/Llama-3.1-8B-Instruct} \\
        Llama 3.1 (70B) & \texttt{meta-llama/Llama-3.1-70B-Instruct} \\
        Llama 3.2 (3B)  & \texttt{meta-llama/Llama-3.2-3B-Instruct} \\
        Llama 3.3 (70B) & \texttt{meta-llama/Llama-3.3-70B-Instruct} \\
        \midrule
        Qwen 2.5 (3B)  & \texttt{Qwen/Qwen2.5-3B-Instruct} \\
        Qwen 2.5 (7B)  & \texttt{Qwen/Qwen2.5-7B-Instruct} \\
        Qwen 2.5 (14B) & \texttt{Qwen/Qwen2.5-14B-Instruct} \\
        Qwen 2.5 (32B) & \texttt{Qwen/Qwen2.5-32B-Instruct} \\
        Qwen 2.5 (72B) & \texttt{Qwen/Qwen2.5-72B-Instruct} \\
        \midrule
        Qwen 3 (0.6B)  & \texttt{Qwen/Qwen3-0.6B} \\
        Qwen 3 (1.7B)  & \texttt{Qwen/Qwen3-1.7B} \\
        Qwen 3 (4B)    & \texttt{Qwen/Qwen3-4B} \\
        Qwen 3 (8B)    & \texttt{Qwen/Qwen3-8B} \\
        Qwen 3 (14B)   & \texttt{Qwen/Qwen3-14B} \\
        Qwen 3 (32B)   & \texttt{Qwen/Qwen3-32B} \\
        \midrule
        Qwen 3 Thinking (0.6B)   & \texttt{Qwen/Qwen3-0.6B} \& \texttt{\# reasoning tokens: 512} \\
        Qwen 3 Thinking (1.7B)   & \texttt{Qwen/Qwen3-1.7B} \& \texttt{\# reasoning tokens: 512}\\
        Qwen 3 Thinking (4B)    & \texttt{Qwen/Qwen3-4B} \& \texttt{\# reasoning tokens: 512}\\
        Qwen 3 Thinking (8B)   & \texttt{Qwen/Qwen3-8B} \& \texttt{\# reasoning tokens: 512}\\
        Qwen 3 Thinking (14B)   & \texttt{Qwen/Qwen3-14B} \& \texttt{\# reasoning tokens: 512}\\
        Qwen 3 Thinking (32B)   & \texttt{Qwen/Qwen3-32B} \& \texttt{\# reasoning tokens: 512}\\
        \midrule
        GPT-5.2             & \texttt{gpt-5.2-2025-12-11} \\
        GPT-5.2 Thinking & \texttt{gpt-5.2-2025-12-11} \& \texttt{reasoning effort: medium} \\
        GPT-5.4             & \texttt{gpt-5.4-2026-03-05} \\
        GPT-5.4 Thinking & \texttt{gpt-5.4-2026-03-05} \& \texttt{reasoning effort: medium} \\
        \bottomrule
    \end{tabular}
    \caption{Model names as referenced in the paper and their corresponding HuggingFace or OpenAI identifiers. We also provide the thinking budget for the models that leverage their reasoning capabilities.}
    \label{tab:model-names}
\end{table*}

\clearpage

\twocolumn

\section{Prompt Templates}

\label{sec:app:prompt_templates}

\definecolor{systempurple}{RGB}{127,119,221}
\definecolor{promptgreen}{RGB}{29,158,117}
\definecolor{lightgray}{RGB}{248,248,248}
\definecolor{boxgray}{RGB}{240,240,240}

\begin{figure}[h!]
\centering
\begin{tcolorbox}[
  colback=white,
  colframe=black!20,
  width=\columnwidth,
  arc=4pt,
  boxrule=0.5pt,
  title={\textbf{Model Prompt}},
  fonttitle=\small\sffamily
]
\textcolor{black!50}{\footnotesize\textsc{System Prompt}}\\[-10pt]
\begin{tcolorbox}[
  colback=lightgray,
  colframe=systempurple,
  leftrule=3pt, toprule=0.3pt, bottomrule=0.3pt, rightrule=0.3pt,
  arc=2pt,
  boxsep=3pt,
  left=6pt, right=6pt, top=4pt, bottom=4pt
]
\small You are an expert in pragmatic inference and in identifying the intended meaning of utterances.
\end{tcolorbox}
\vspace{1pt}
\textcolor{black!50}{\footnotesize\textsc{Prompt Template}}\\[-10pt]
\begin{tcolorbox}[
  colback=lightgray,
  colframe=promptgreen,
  leftrule=3pt, toprule=0.3pt, bottomrule=0.3pt, rightrule=0.3pt,
  arc=2pt,
  boxsep=3pt,
  left=6pt, right=6pt, top=4pt, bottom=4pt
]
\small You will be given an utterance as well as a potential interpretation for this utterance. Your task is to decide whether the interpretation provided for the utterance is true or false. You will be given a numbered list containing true and false. Read the utterance and the potential interpretation and pick the number corresponding to whether the interpretation is true or false. 
\colorbox{systempurple!15}{\texttt{\{scenario\}}}
Interpretation: \colorbox{promptgreen!15}{\texttt{\{implicature\}}}
\colorbox{orange!15}{\texttt{\{question\}}}
\colorbox{blue!10}{\texttt{\{options\}}}
Answer:
\end{tcolorbox}
\end{tcolorbox}
\caption{Model prompt used for pragmatic inference evaluation. Variables
\colorbox{systempurple!15}{\texttt{\{scenario\}}},
\colorbox{promptgreen!15}{\texttt{\{implicature\}}},
\colorbox{orange!15}{\texttt{\{question\}}}, and
\colorbox{blue!10}{\texttt{\{options\}}} are filled per instance.}
\label{fig:prompt-template}
\end{figure}
\newpage
\begin{figure}[h!]
\centering
\begin{tcolorbox}[
  colback=white,
  colframe=black!20,
  width=\columnwidth,
  arc=4pt,
  boxrule=0.5pt,
  title={\textbf{Example Instance}},
  fonttitle=\small\sffamily
]
\begin{tcolorbox}[
  colback=white,
  colframe=black!15,
  arc=4pt,
  boxrule=0.5pt,
  boxsep=0pt,
  left=0pt, right=0pt, top=0pt, bottom=0pt
]
  \begin{tabular}{@{\hspace{8pt}} p{0.08\linewidth} p{0.80\linewidth} @{}}
    \small $\trig$   & \small and some places, they, they, they really nail them for tax. \\[4pt]
    \small $\belief$ & \small Some, but not all, places nail them for tax. \\[4pt]
    \small $\cancel$ & \small i think they all do that now \\
  \end{tabular}
  \begin{tcolorbox}[
    colback=boxgray,
    colframe=black!0,
    arc=0pt,
    boxrule=0pt,
    left=8pt, right=8pt, top=6pt, bottom=6pt
  ]
  \textcolor{black!50}{\footnotesize\textsc{Filled Prompt (without $\cancel$)}}\\[4pt]
  \small
  You will be given an utterance as well as a potential interpretation
  for this utterance. Your task is to decide whether the interpretation
  provided for the utterance is true or false. You will be given a
  numbered list containing true and false. Read the utterance and the
  potential interpretation and pick the number corresponding to whether
  the interpretation is true or false.

  \medskip
  Utterance: and some places, they, they, they really nail them for tax.

  \medskip
  Interpretation: Some, but not all, places nail them for tax.

  \medskip
  Is the interpretation of the previous utterance true or false?\\
  1: True\\
  2: False

  \medskip
  Answer:
  \end{tcolorbox}

  \begin{tcolorbox}[
    colback=boxgray,
    colframe=black!0,
    arc=0pt,
    boxrule=0pt,
    left=8pt, right=8pt, top=6pt, bottom=6pt
  ]
  \textcolor{black!50}{\footnotesize\textsc{Filled Prompt (with $\cancel$)}}\\[4pt]
  \small
  You will be given an utterance as well as a potential interpretation
  for this utterance. Your task is to decide whether the interpretation
  provided for the utterance is true or false. You will be given a
  numbered list containing true and false. Read the utterance and the
  potential interpretation and pick the number corresponding to whether
  the interpretation is true or false.

  \medskip
  Utterance: and some places, they, they, they really nail them for tax.
  i think they all do that now

  \medskip
  Interpretation: Some, but not all, places nail them for tax.

  \medskip
  Is the interpretation of the previous utterance true or false?\\
  1: True\\
  2: False

  \medskip
  Answer:
  \end{tcolorbox}
\end{tcolorbox}
\end{tcolorbox}
\caption{Example instance for the prompt in Figure~\ref{fig:prompt-template},
shown without (top) and with (bottom) the cancellation $\cancel$ appended to
the utterance in \texttt{\{scenario\}}. }
\label{fig:prompt-example}
\end{figure}
\clearpage

\section{Control Datasets}

\label{sec:app:control_datasets}

To conduct our control experiments in Section~\ref{sec:cancel_experiments}, we created three additional datasets: \negdataset, \strengthdataset and \neutraldataset. All of these datasets have the same size as the original \datasetname dataset. They share the same context, triggering utterance and implicature as the original dataset but differ in their follow-up utterance (i.e., the utterance which follows the cancelling utterance). We provide details regarding how their follow-up utterances were created.

The \negdataset follow-up utterance, $\explicit$, is a cancelling utterance with a discourse marker and an explicit negation of the implicature. We first determined which discourse markers best preserved coherence in each of the dataset splits and then applied formulaic negations to the implicature. For instance, to explicitly negate a scalar implicature like ``...some of them showed up.'' we use discourse markers like ``in fact'', ``actually'' or ``as a matter of fact'' followed by the negation of the scalar implicature ``all of them showed up.'' On the other hand, conversational implicatures suffer in coherence when negated using the same discourse markers. Thus, in this case, we use the discourse marker ``Though, I don't mean to imply that'' construction. For instance, ``I'm more behind than you. Though, I don't mean to imply that I can't help you with this problem.''

The \strengthdataset follow-up utterance, $\strength$, is a strengthening utterance: an utterance designed to confirm the implicature triggered by the triggering utterance. We manually created these utterances while controlling for coherence. For instance, to strengthen a scalar implicature, we reinforce the fact that not the entire set of elements being discussed should be included e.g., ``and some places, they, they, they really nail them for tax. though some get that exemption i was talking about''. For the other types of implicatures, we create similar utterances which implicitly reinforce the implicature. For instance, in the case of synthetic conversational implicatures, the implicature ``I cannot help you with this question.'' can be reinforced with the utterance ``Is there no one else you can ask?''

The \neutraldataset follow-up utterance, $\neutral$, is a neutral utterance: an utterance which is irrelevant to the triggering utterance and the implicature. The neutral utterance is selected by randomly sampling a cancelling utterance from the same corresponding dataset split.

\section{Additional Results}
\label{sec:app:results}

\paragraph{Full Results}

\begin{figure}[p]
    \centering
    \includegraphics[width=\linewidth]{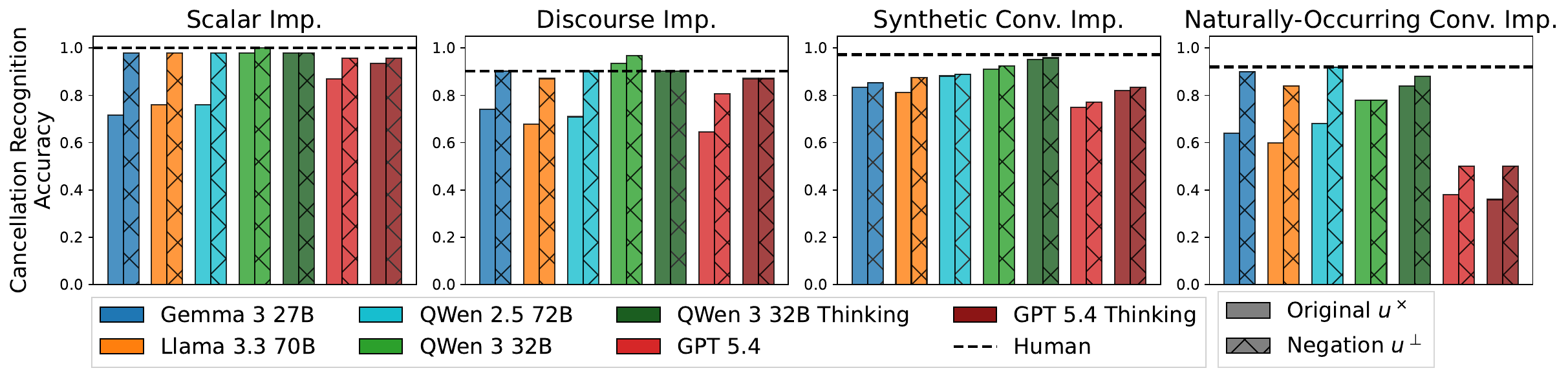}
    \caption{Cancellation recognition accuracy using original cancelling utterances from \datasetname and the explicit cancelling utterances from \negdataset which contain negation.}
    \label{fig:app:belief_update_infact}
\end{figure}

\begin{figure}[p]
    \centering
    \includegraphics[width=\linewidth]{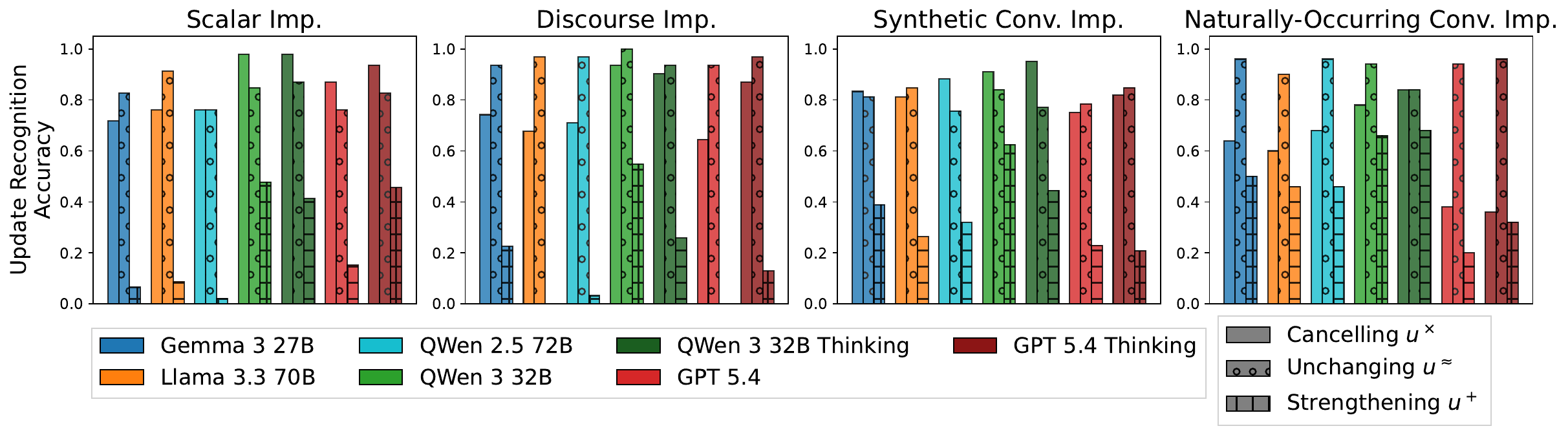}
    \caption{Update belief accuracies given different update types. Cancelling denotes the update belief as triggered by implicature cancellation while unchanging denotes the belief update accuracy using \neutraldataset items and strengthening denotes the belief update accuracy using \strengthdataset items.
    }
    \label{fig:app:belief_update_by_type}
\end{figure}

We provide full implicature recognition, cancellation recognition and belief update accuracies in Tables~\ref{tab:implicature_acc},~\ref{tab:decrease_acc}~and~\ref{tab:impli_decrease_acc} respectively. We also include the full results for our control experiments in Tables~\ref{tab:implicature_acc_on_prior},~\ref{tab:cancellation_acc_on_bot},~\ref{tab:strengthening_acc},~\ref{tab:neutral_acc_approx}. In addition, the form and update type controls on all the implicature types are plotted in Figures~\ref{fig:app:belief_update_infact}~and~\ref{fig:app:belief_update_by_type}. 

\paragraph{Does length explain implicature and cancellation recognition difficulty?}

To determine whether our results are confounded by length and the ability of LLMs to perform under longer prompts, we run a correlation analysis between different length variables and implicature and cancellation recognition probabilities. In particular, we compute Kendall tau's concordances between different space-separated lengths, $|\cdot| : \mathcal{V}^* \rightarrow \mathbb{N}$, and the implicature and cancellation probabilities of Llama 3.3 70B, the model which performed the best overall in this study. We provide Kendall tau's concordances $\tau$ between $\probimp$ and the space-separated lengths of $\context$, $\trig$ and $\belief$ in Table~\ref{tab:kendall_impli}. We also provide Kendall tau's concordances $\tau$ between $\probcancel$ and the space-separated lengths of $\context$, $\trig$, $\belief$ and $\cancel$ in Table~\ref{tab:kendall_cancel}.

Our results in Tables~\ref{tab:kendall_impli}~and~\ref{tab:kendall_cancel} indicate weak concordances between the tested length variables and the LLM-induced probabilities $\probimp$ and $\probcancel$. In addition, in most cases, the computed concordances are not statistically significant. The only statistically significant Kendall's tau values between \probimp and the length variables are $0.27$ (context length of the naturally-occurring implicatures) and $0.28$ (utterance length of the scalar implicatures). The only statistically significant Kendall's tau between \probcancel and the length variables is $0.22$ (triggering utterance length of the scalar implicatures). All other concordances are near zero and not statistically significant. While a lack of concordance is not evidence that there is no relation between length and implicature and cancellation difficulty, we believe it suggests that the difficulty of the items in \datasetname cannot be explained by length alone.

\onecolumn

\begin{table}[htbp]
\centering
\resizebox{\textwidth}{!}{%
\begin{tabular}{lcccc}
\toprule
 Model & \makecell{Scalar\\Implicature} & \makecell{Discourse\\Implicature} & \makecell{Synthetic\\Conversational Implicature} & \makecell{Naturally-Occurring\\Conversational Implicature} \\
\midrule
 \textbf{Gemma 3}                    &  &  &  &  \\
 Gemma 3 (4B)                        & 1.000 & 0.871 & 0.708 & 0.700 \\
 Gemma 3 (12B)                       & 1.000 & 0.903 & 0.722 & 0.560 \\
 Gemma 3 (27B)                       & 1.000 & 0.806 & 0.674 & 0.460 \\
\midrule
 \textbf{Llama 3}                    &  &  &  &  \\
 Llama 3.1 (8B)                      & 0.913 & 0.677 & 0.229 & 0.160 \\
 Llama 3.1 (70B)                     & 0.978 & 0.935 & 0.806 & 0.560 \\
 Llama 3.2 (3B)                      & 0.478 & 0.452 & 0.271 & 0.100 \\
 Llama 3.3 (70B)                     & 0.957 & 0.935 & 0.771 & 0.600 \\
\midrule
 \textbf{Qwen 2.5}                   &  &  &  &  \\
 Qwen 2.5 (3B)                       & 0.522 & 0.419 & 0.076 & 0.080 \\
 Qwen 2.5 (7B)                       & 0.761 & 0.613 & 0.222 & 0.160 \\
 Qwen 2.5 (14B)                      & 0.891 & 0.806 & 0.479 & 0.340 \\
 Qwen 2.5 (32B)                      & 0.935 & 0.871 & 0.556 & 0.380 \\
 Qwen 2.5 (72B)                      & 1.000 & 0.871 & 0.694 & 0.440 \\
\midrule
 \textbf{Qwen 3}                     &  &  &  &  \\
 Qwen 3 (0.6B)                       & 0.000 & 0.000 & 0.000 & 0.000 \\
 Qwen 3 (1.7B)                       & 0.848 & 0.871 & 0.715 & 0.600 \\
 Qwen 3 (4B)                         & 1.000 & 0.903 & 0.590 & 0.780 \\
 Qwen 3 (8B)                         & 1.000 & 0.839 & 0.326 & 0.420 \\
 Qwen 3 (14B)                        & 0.891 & 0.903 & 0.674 & 0.420 \\
 Qwen 3 (32B)                        & 0.957 & 0.871 & 0.708 & 0.480 \\
\midrule
 \textbf{Qwen 3 Thinking}            &  &  &  &  \\
 Qwen 3 (0.6B)                       & 0.978 & 0.613 & 0.444 & 0.500 \\
 Qwen 3 (1.7B)                       & 0.435 & 0.774 & 0.292 & 0.400 \\
 Qwen 3 (4B)                         & 1.000 & 0.839 & 0.521 & 0.460 \\
 Qwen 3 (8B)                         & 0.804 & 0.742 & 0.347 & 0.380 \\
 Qwen 3 (14B)                        & 0.587 & 0.839 & 0.528 & 0.420 \\
 Qwen 3 (32B)                        & 0.935 & 0.903 & 0.771 & 0.420 \\
\midrule
 \textbf{GPT}                        &  &  &  &  \\
 GPT 5.2                             & 0.804 & 0.806 & 0.674 & 0.500 \\
 GPT 5.4                             & 0.891 & 0.903 & 0.764 & 0.440 \\
\midrule
 \textbf{GPT Thinking}               &  &  &  &  \\
 GPT 5.2                             & 0.674 & 0.871 & 0.743 & 0.460 \\
 GPT 5.4                             & 0.826 & 0.839 & 0.806 & 0.440 \\
\midrule
 \textbf{Human (avg.)}               & 1.000 & 0.903 & 0.910 & 0.780 \\
\bottomrule
\end{tabular}%
}
\caption{Full implicature recognition accuracy results.}
\label{tab:implicature_acc}
\end{table}

\begin{table}[htbp]
\centering
\resizebox{\textwidth}{!}{%
\begin{tabular}{lcccc}
\toprule
 Model & \makecell{Scalar\\Implicature} & \makecell{Discourse\\Implicature} & \makecell{Synthetic\\Conversational Implicature} & \makecell{Naturally-Occurring\\Conversational Implicature} \\
\midrule
 \textbf{Gemma 3}                    &  &  &  &  \\
 Gemma 3 (4B)                        & 0.565 & 0.710 & 0.826 & 0.720 \\
 Gemma 3 (12B)                       & 0.326 & 0.548 & 0.743 & 0.560 \\
 Gemma 3 (27B)                       & 0.717 & 0.742 & 0.833 & 0.640 \\
\midrule
 \textbf{Llama 3}                    &  &  &  &  \\
 Llama 3.1 (8B)                      & 0.870 & 0.839 & 0.944 & 0.660 \\
 Llama 3.1 (70B)                     & 0.913 & 0.774 & 0.944 & 0.820 \\
 Llama 3.2 (3B)                      & 0.848 & 0.806 & 0.729 & 0.540 \\
 Llama 3.3 (70B)                     & 0.761 & 0.677 & 0.812 & 0.600 \\
\midrule
 \textbf{Qwen 2.5}                   &  &  &  &  \\
 Qwen 2.5 (3B)                       & 0.674 & 0.774 & 0.438 & 0.560 \\
 Qwen 2.5 (7B)                       & 0.891 & 0.710 & 0.819 & 0.620 \\
 Qwen 2.5 (14B)                      & 0.826 & 0.677 & 0.847 & 0.720 \\
 Qwen 2.5 (32B)                      & 0.870 & 0.710 & 0.917 & 0.740 \\
 Qwen 2.5 (72B)                      & 0.761 & 0.710 & 0.882 & 0.680 \\
\midrule
 \textbf{Qwen 3}                     &  &  &  &  \\
 Qwen 3 (0.6B)                       & 0.457 & 0.742 & 0.556 & 0.520 \\
 Qwen 3 (1.7B)                       & 0.609 & 0.516 & 0.688 & 0.680 \\
 Qwen 3 (4B)                         & 0.304 & 0.645 & 0.875 & 0.660 \\
 Qwen 3 (8B)                         & 0.826 & 0.710 & 0.840 & 0.640 \\
 Qwen 3 (14B)                        & 0.674 & 0.613 & 0.750 & 0.640 \\
 Qwen 3 (32B)                        & 0.978 & 0.935 & 0.910 & 0.780 \\
\midrule
 \textbf{Qwen 3 Thinking}            &  &  &  &  \\
 Qwen 3 (0.6B)                       & 0.826 & 0.742 & 0.660 & 0.680 \\
 Qwen 3 (1.7B)                       & 0.870 & 0.613 & 0.736 & 0.600 \\
 Qwen 3 (4B)                         & 0.826 & 0.710 & 0.868 & 0.760 \\
 Qwen 3 (8B)                         & 0.957 & 0.871 & 0.938 & 0.740 \\
 Qwen 3 (14B)                        & 0.978 & 0.871 & 0.931 & 0.880 \\
 Qwen 3 (32B)                        & 0.978 & 0.903 & 0.951 & 0.840 \\
\midrule
 \textbf{GPT}                        &  &  &  &  \\
 GPT 5.2                             & 0.891 & 0.613 & 0.715 & 0.320 \\
 GPT 5.4                             & 0.870 & 0.645 & 0.750 & 0.380 \\
\midrule
 \textbf{GPT Thinking}               &  &  &  &  \\
 GPT 5.2                             & 0.913 & 0.871 & 0.785 & 0.460 \\
 GPT 5.4                             & 0.935 & 0.871 & 0.819 & 0.360 \\
\midrule
 \textbf{Human (avg.)}               & 1.000 & 0.903 & 0.972 & 0.920 \\
\bottomrule
\end{tabular}%
}
\caption{Full cancellation recognition accuracy results.}
\label{tab:decrease_acc}
\end{table}

\begin{table}[htbp]
\centering
\resizebox{\textwidth}{!}{%
\begin{tabular}{lcccc}
\toprule
 Model & \makecell{Scalar\\Implicature} & \makecell{Discourse\\Implicature} & \makecell{Synthetic\\Conversational Implicature} & \makecell{Naturally-Occurring\\Conversational Implicature} \\
\midrule
 \textbf{Gemma 3}                    &  &  &  &  \\
 Gemma 3 (4B)                        & 0.565 & 0.645 & 0.590 & 0.540 \\
 Gemma 3 (12B)                       & 0.326 & 0.484 & 0.576 & 0.280 \\
 Gemma 3 (27B)                       & 0.717 & 0.581 & 0.569 & 0.240 \\
\midrule
 \textbf{Llama 3}                    &  &  &  &  \\
 Llama 3.1 (8B)                      & 0.804 & 0.645 & 0.222 & 0.160 \\
 Llama 3.1 (70B)                     & 0.891 & 0.742 & 0.792 & 0.460 \\
 Llama 3.2 (3B)                      & 0.457 & 0.452 & 0.271 & 0.100 \\
 Llama 3.3 (70B)                     & 0.717 & 0.613 & 0.660 & 0.360 \\
\midrule
 \textbf{Qwen 2.5}                   &  &  &  &  \\
 Qwen 2.5 (3B)                       & 0.370 & 0.387 & 0.069 & 0.040 \\
 Qwen 2.5 (7B)                       & 0.674 & 0.484 & 0.215 & 0.140 \\
 Qwen 2.5 (14B)                      & 0.717 & 0.484 & 0.438 & 0.200 \\
 Qwen 2.5 (32B)                      & 0.804 & 0.581 & 0.521 & 0.280 \\
 Qwen 2.5 (72B)                      & 0.761 & 0.613 & 0.611 & 0.240 \\
\midrule
 \textbf{Qwen 3}                     &  &  &  &  \\
 Qwen 3 (0.6B)                       & 0.000 & 0.000 & 0.000 & 0.000 \\
 Qwen 3 (1.7B)                       & 0.565 & 0.484 & 0.535 & 0.440 \\
 Qwen 3 (4B)                         & 0.304 & 0.581 & 0.542 & 0.480 \\
 Qwen 3 (8B)                         & 0.826 & 0.613 & 0.292 & 0.240 \\
 Qwen 3 (14B)                        & 0.609 & 0.548 & 0.604 & 0.340 \\
 Qwen 3 (32B)                        & 0.935 & 0.871 & 0.701 & 0.400 \\
\midrule
 \textbf{Qwen 3 Thinking}            &  &  &  &  \\
 Qwen 3 (0.6B)                       & 0.804 & 0.581 & 0.403 & 0.380 \\
 Qwen 3 (1.7B)                       & 0.413 & 0.516 & 0.257 & 0.320 \\
 Qwen 3 (4B)                         & 0.826 & 0.613 & 0.479 & 0.360 \\
 Qwen 3 (8B)                         & 0.761 & 0.677 & 0.340 & 0.280 \\
 Qwen 3 (14B)                        & 0.587 & 0.774 & 0.507 & 0.380 \\
 Qwen 3 (32B)                        & 0.913 & 0.839 & 0.764 & 0.400 \\
\midrule
 \textbf{GPT}                        &  &  &  &  \\
 GPT 5.2                             & 0.804 & 0.484 & 0.646 & 0.300 \\
 GPT 5.4                             & 0.826 & 0.645 & 0.708 & 0.280 \\
\midrule
 \textbf{GPT Thinking}               &  &  &  &  \\
 GPT 5.2                             & 0.652 & 0.806 & 0.715 & 0.400 \\
 GPT 5.4                             & 0.804 & 0.806 & 0.764 & 0.340 \\
\midrule
 \textbf{Human (avg.)}               & 1.000 & 0.903 & 0.903 & 0.720 \\
\bottomrule
\end{tabular}%
}
\caption{Full belief update accuracy results.}
\label{tab:impli_decrease_acc}
\end{table}

\begin{table}[htbp]
\centering
\resizebox{\textwidth}{!}{%
\begin{tabular}{lcccc}
\toprule
 Model & \makecell{Scalar\\Implicature} & \makecell{Discourse\\Implicature} & \makecell{Synthetic\\Conversational Implicature} & \makecell{Naturally-Occurring\\Conversational Implicature} \\
\midrule
 \textbf{Gemma 3}                    &  &  &  &  \\
 Gemma 3 (4B)                        & 1.000 & 0.774 & 1.000 & 0.920 \\
 Gemma 3 (12B)                       & 1.000 & 0.806 & 0.965 & 0.660 \\
 Gemma 3 (27B)                       & 0.957 & 0.613 & 0.951 & 0.560 \\
\midrule
 \textbf{Llama 3}                    &  &  &  &  \\
 Llama 3.1 (8B)                      & 1.000 & 0.677 & 0.854 & 0.620 \\
 Llama 3.1 (70B)                     & 0.978 & 0.710 & 0.868 & 0.580 \\
 Llama 3.2 (3B)                      & 1.000 & 0.806 & 0.979 & 0.820 \\
 Llama 3.3 (70B)                     & 1.000 & 0.710 & 0.882 & 0.640 \\
\midrule
 \textbf{Qwen 2.5}                   &  &  &  &  \\
 Qwen 2.5 (3B)                       & 0.087 & 0.032 & 0.021 & 0.020 \\
 Qwen 2.5 (7B)                       & 0.957 & 0.194 & 0.465 & 0.200 \\
 Qwen 2.5 (14B)                      & 0.891 & 0.290 & 0.167 & 0.160 \\
 Qwen 2.5 (32B)                      & 0.957 & 0.484 & 0.806 & 0.420 \\
 Qwen 2.5 (72B)                      & 1.000 & 0.677 & 0.833 & 0.460 \\
\midrule
 \textbf{Qwen 3}                     &  &  &  &  \\
 Qwen 3 (0.6B)                       & 0.000 & 0.000 & 0.000 & 0.000 \\
 Qwen 3 (1.7B)                       & 1.000 & 0.742 & 0.910 & 0.620 \\
 Qwen 3 (4B)                         & 1.000 & 0.710 & 0.910 & 0.580 \\
 Qwen 3 (8B)                         & 0.978 & 0.452 & 0.715 & 0.380 \\
 Qwen 3 (14B)                        & 1.000 & 0.323 & 0.549 & 0.260 \\
 Qwen 3 (32B)                        & 0.935 & 0.516 & 0.729 & 0.340 \\
\midrule
 \textbf{Qwen 3 Thinking}            &  &  &  &  \\
 Qwen 3 (0.6B)                       & 0.913 & 0.355 & 0.611 & 0.380 \\
 Qwen 3 (1.7B)                       & 0.500 & 0.323 & 0.292 & 0.100 \\
 Qwen 3 (4B)                         & 0.935 & 0.419 & 0.396 & 0.140 \\
 Qwen 3 (8B)                         & 0.913 & 0.258 & 0.486 & 0.180 \\
 Qwen 3 (14B)                        & 0.935 & 0.258 & 0.576 & 0.260 \\
 Qwen 3 (32B)                        & 0.957 & 0.548 & 0.757 & 0.420 \\
\midrule
 \textbf{GPT}                        &  &  &  &  \\
 GPT 5.2                             & 0.978 & 0.290 & 0.389 & 0.260 \\
 GPT 5.4                             & 0.935 & 0.387 & 0.500 & 0.260 \\
\midrule
 \textbf{GPT Thinking}               &  &  &  &  \\
 GPT 5.2                             & 0.935 & 0.323 & 0.340 & 0.220 \\
 GPT 5.4                             & 0.957 & 0.452 & 0.444 & 0.200 \\
\midrule
 \textbf{Human (avg.)}               & 1.000 & 0.903 & 0.910 & 0.780 \\
\bottomrule
\end{tabular}%
}
\caption{Full implicature recognition accuracy results for the prior common ground control experiment.}
\label{tab:implicature_acc_on_prior}
\end{table}

\begin{table}[htbp]
\centering
\resizebox{\textwidth}{!}{%
\begin{tabular}{lcccc}
\toprule
 Model & \makecell{Scalar\\Implicature} & \makecell{Discourse\\Implicature} & \makecell{Synthetic\\Conversational Implicature} & \makecell{Naturally-Occurring\\Conversational Implicature} \\
\midrule
 \textbf{Gemma 3}                    &  &  &  &  \\
 Gemma 3 (4B)                        & 0.891 & 0.968 & 0.812 & 0.820 \\
 Gemma 3 (12B)                       & 0.804 & 0.871 & 0.812 & 0.760 \\
 Gemma 3 (27B)                       & 0.978 & 0.903 & 0.854 & 0.900 \\
\midrule
 \textbf{Llama 3}                    &  &  &  &  \\
 Llama 3.1 (8B)                      & 1.000 & 1.000 & 0.938 & 0.840 \\
 Llama 3.1 (70B)                     & 0.978 & 0.935 & 0.951 & 0.940 \\
 Llama 3.2 (3B)                      & 0.978 & 1.000 & 0.833 & 0.840 \\
 Llama 3.3 (70B)                     & 0.978 & 0.871 & 0.875 & 0.840 \\
\midrule
 \textbf{Qwen 2.5}                   &  &  &  &  \\
 Qwen 2.5 (3B)                       & 0.957 & 0.806 & 0.479 & 0.540 \\
 Qwen 2.5 (7B)                       & 1.000 & 0.903 & 0.868 & 0.820 \\
 Qwen 2.5 (14B)                      & 0.957 & 0.871 & 0.889 & 0.900 \\
 Qwen 2.5 (32B)                      & 1.000 & 0.871 & 0.965 & 0.940 \\
 Qwen 2.5 (72B)                      & 0.978 & 0.903 & 0.889 & 0.920 \\
\midrule
 \textbf{Qwen 3}                     &  &  &  &  \\
 Qwen 3 (0.6B)                       & 0.413 & 0.806 & 0.542 & 0.500 \\
 Qwen 3 (1.7B)                       & 0.913 & 0.871 & 0.701 & 0.820 \\
 Qwen 3 (4B)                         & 0.870 & 0.968 & 0.833 & 0.880 \\
 Qwen 3 (8B)                         & 1.000 & 0.968 & 0.847 & 0.900 \\
 Qwen 3 (14B)                        & 0.978 & 0.871 & 0.812 & 0.680 \\
 Qwen 3 (32B)                        & 1.000 & 0.968 & 0.924 & 0.780 \\
\midrule
 \textbf{Qwen 3 Thinking}            &  &  &  &  \\
 Qwen 3 (0.6B)                       & 0.957 & 0.839 & 0.792 & 0.740 \\
 Qwen 3 (1.7B)                       & 0.978 & 0.871 & 0.785 & 0.880 \\
 Qwen 3 (4B)                         & 0.978 & 0.968 & 0.882 & 0.960 \\
 Qwen 3 (8B)                         & 1.000 & 0.968 & 0.910 & 0.940 \\
 Qwen 3 (14B)                        & 0.978 & 0.935 & 0.924 & 0.920 \\
 Qwen 3 (32B)                        & 0.978 & 0.903 & 0.958 & 0.880 \\
\midrule
 \textbf{GPT}                        &  &  &  &  \\
 GPT 5.2                             & 0.891 & 0.839 & 0.750 & 0.480 \\
 GPT 5.4                             & 0.957 & 0.806 & 0.771 & 0.500 \\
\midrule
 \textbf{GPT Thinking}               &  &  &  &  \\
 GPT 5.2                             & 0.935 & 0.839 & 0.799 & 0.500 \\
 GPT 5.4                             & 0.957 & 0.871 & 0.833 & 0.500 \\
\midrule
 \textbf{Human (avg.)}               & 1.000 & 0.903 & 0.972 & 0.920 \\
\bottomrule
\end{tabular}%
}
\caption{Full cancellation recognition accuracy results on the \negdataset items.}
\label{tab:cancellation_acc_on_bot}
\end{table}

\begin{table}[htbp]
\centering
\resizebox{\textwidth}{!}{%
\begin{tabular}{lcccc}
\toprule
 Model & \makecell{Scalar\\Implicature} & \makecell{Discourse\\Implicature} & \makecell{Synthetic\\Conversational Implicature} & \makecell{Naturally-Occurring\\Conversational Implicature} \\
\midrule
 \textbf{Gemma 3}                    &  &  &  &  \\
 Gemma 3 (4B)                        & 0.174 & 0.226 & 0.424 & 0.620 \\
 Gemma 3 (12B)                       & 0.022 & 0.129 & 0.382 & 0.540 \\
 Gemma 3 (27B)                       & 0.065 & 0.226 & 0.389 & 0.500 \\
\midrule
 \textbf{Llama 3}                    &  &  &  &  \\
 Llama 3.1 (8B)                      & 0.522 & 0.484 & 0.535 & 0.680 \\
 Llama 3.1 (70B)                     & 0.261 & 0.161 & 0.528 & 0.600 \\
 Llama 3.2 (3B)                      & 0.522 & 0.613 & 0.458 & 0.580 \\
 Llama 3.3 (70B)                     & 0.087 & 0.000 & 0.264 & 0.460 \\
\midrule
 \textbf{Qwen 2.5}                   &  &  &  &  \\
 Qwen 2.5 (3B)                       & 0.761 & 0.226 & 0.604 & 0.600 \\
 Qwen 2.5 (7B)                       & 0.565 & 0.226 & 0.674 & 0.560 \\
 Qwen 2.5 (14B)                      & 0.196 & 0.097 & 0.521 & 0.520 \\
 Qwen 2.5 (32B)                      & 0.130 & 0.161 & 0.465 & 0.500 \\
 Qwen 2.5 (72B)                      & 0.022 & 0.032 & 0.319 & 0.460 \\
\midrule
 \textbf{Qwen 3}                     &  &  &  &  \\
 Qwen 3 (0.6B)                       & 0.478 & 0.419 & 0.444 & 0.440 \\
 Qwen 3 (1.7B)                       & 0.630 & 0.226 & 0.521 & 0.380 \\
 Qwen 3 (4B)                         & 0.000 & 0.065 & 0.583 & 0.520 \\
 Qwen 3 (8B)                         & 0.196 & 0.194 & 0.625 & 0.620 \\
 Qwen 3 (14B)                        & 0.217 & 0.129 & 0.653 & 0.620 \\
 Qwen 3 (32B)                        & 0.478 & 0.548 & 0.625 & 0.660 \\
\midrule
 \textbf{Qwen 3 Thinking}            &  &  &  &  \\
 Qwen 3 (0.6B)                       & 0.370 & 0.484 & 0.493 & 0.580 \\
 Qwen 3 (1.7B)                       & 0.804 & 0.194 & 0.528 & 0.680 \\
 Qwen 3 (4B)                         & 0.065 & 0.097 & 0.444 & 0.380 \\
 Qwen 3 (8B)                         & 0.674 & 0.290 & 0.576 & 0.640 \\
 Qwen 3 (14B)                        & 0.674 & 0.323 & 0.528 & 0.580 \\
 Qwen 3 (32B)                        & 0.413 & 0.258 & 0.444 & 0.680 \\
\midrule
 \textbf{GPT}                        &  &  &  &  \\
 GPT 5.2                             & 0.261 & 0.065 & 0.319 & 0.140 \\
 GPT 5.4                             & 0.152 & 0.000 & 0.229 & 0.200 \\
\midrule
 \textbf{GPT Thinking}               &  &  &  &  \\
 GPT 5.2                             & 0.630 & 0.097 & 0.306 & 0.300 \\
 GPT 5.4                             & 0.457 & 0.129 & 0.208 & 0.320 \\
\bottomrule
\end{tabular}%
}
\caption{Full belief strengthening accuracy results on the \strengthdataset items.}
\label{tab:strengthening_acc}
\end{table}

\begin{table}[htbp]
\centering
\resizebox{\textwidth}{!}{%
\begin{tabular}{lcccc}
\toprule
 Model & \makecell{Scalar\\Implicature} & \makecell{Discourse\\Implicature} & \makecell{Synthetic\\Conversational Implicature} & \makecell{Naturally-Occurring\\Conversational Implicature} \\
\midrule
 \textbf{Gemma 3}                    &  &  &  &  \\
 Gemma 3 (4B)                        & 1.000 & 1.000 & 0.757 & 0.860 \\
 Gemma 3 (12B)                       & 0.913 & 0.968 & 0.792 & 0.940 \\
 Gemma 3 (27B)                       & 0.826 & 0.935 & 0.812 & 0.960 \\
\midrule
 \textbf{Llama 3}                    &  &  &  &  \\
 Llama 3.1 (8B)                      & 0.674 & 0.806 & 0.868 & 0.900 \\
 Llama 3.1 (70B)                     & 0.913 & 0.935 & 0.847 & 0.900 \\
 Llama 3.2 (3B)                      & 0.609 & 0.742 & 0.819 & 0.940 \\
 Llama 3.3 (70B)                     & 0.913 & 0.968 & 0.847 & 0.900 \\
\midrule
 \textbf{Qwen 2.5}                   &  &  &  &  \\
 Qwen 2.5 (3B)                       & 0.761 & 0.806 & 0.931 & 0.920 \\
 Qwen 2.5 (7B)                       & 0.674 & 0.871 & 0.868 & 0.980 \\
 Qwen 2.5 (14B)                      & 0.630 & 0.806 & 0.757 & 0.900 \\
 Qwen 2.5 (32B)                      & 0.587 & 0.903 & 0.792 & 0.920 \\
 Qwen 2.5 (72B)                      & 0.761 & 0.968 & 0.757 & 0.960 \\
\midrule
 \textbf{Qwen 3}                     &  &  &  &  \\
 Qwen 3 (0.6B)                       & 1.000 & 1.000 & 1.000 & 1.000 \\
 Qwen 3 (1.7B)                       & 0.783 & 0.935 & 0.847 & 0.900 \\
 Qwen 3 (4B)                         & 1.000 & 0.968 & 0.826 & 0.920 \\
 Qwen 3 (8B)                         & 0.826 & 0.903 & 0.826 & 0.940 \\
 Qwen 3 (14B)                        & 0.870 & 1.000 & 0.799 & 0.940 \\
 Qwen 3 (32B)                        & 0.848 & 1.000 & 0.840 & 0.940 \\
\midrule
 \textbf{Qwen 3 Thinking}            &  &  &  &  \\
 Qwen 3 (0.6B)                       & 0.870 & 0.774 & 0.674 & 0.660 \\
 Qwen 3 (1.7B)                       & 0.457 & 0.903 & 0.771 & 0.840 \\
 Qwen 3 (4B)                         & 1.000 & 0.903 & 0.806 & 0.880 \\
 Qwen 3 (8B)                         & 0.739 & 0.871 & 0.771 & 0.940 \\
 Qwen 3 (14B)                        & 0.674 & 0.903 & 0.743 & 0.920 \\
 Qwen 3 (32B)                        & 0.870 & 0.935 & 0.771 & 0.840 \\
\midrule
 \textbf{GPT}                        &  &  &  &  \\
 GPT 5.2                             & 0.696 & 0.968 & 0.812 & 0.940 \\
 GPT 5.4                             & 0.761 & 0.935 & 0.785 & 0.940 \\
\midrule
 \textbf{GPT Thinking}               &  &  &  &  \\
 GPT 5.2                             & 0.717 & 0.871 & 0.847 & 0.960 \\
 GPT 5.4                             & 0.826 & 0.968 & 0.847 & 0.960 \\
\bottomrule
\end{tabular}%
}
\caption{Full belief unchanging accuracy results on the \neutraldataset items.}
\label{tab:neutral_acc_approx}
\end{table}

\newpage

\clearpage

\begin{table}[p]
\centering
\begin{tabular}{lcccc}
\toprule
 & Scalar Imp. & Discourse Imp. & Synthetic Conv. Imp. & Naturally-Occurring Conv. Imp. \\
\midrule
$|\context|$   & --- & $-0.06$ & $-0.01$ & $0.27^*$ \\
$|\belief|$   & $-0.01$ & $-0.26$ & $0.07$ & $-0.10$ \\
$|\trig|$   & $0.28^*$ & $-0.13$ & $0.06$ & $-0.01$ \\
\bottomrule
\end{tabular}
\caption{Kendall's tau $\tau$ between $\probimp$ and length variables for the dataset splits. Values with $^*$ indicate statistical significance ($p < 0.05$).}
\label{tab:kendall_impli}
\end{table}

\begin{table}[p]
\centering
\begin{tabular}{lccccc}
\toprule
 & Scalar Imp. & Discourse Imp. & Synthetic Conv. Imp. & Naturally-Occurring Conv. Imp. \\
\midrule
$|\context|$   & --- & $-0.21$ & $0.02$ & $0.16$ \\
$|\belief|$   & $0.08$ & $-0.00$ & $0.07$ & $0.02$ \\
$|\trig|$   & $0.22^*$ & $-0.06$ & $0.01$ & $0.14$ \\
$|\cancel|$   & $0.09$ & $-0.05$ & $-0.07$ & $-0.10$ \\
\bottomrule
\end{tabular}
\caption{Kendall's tau $\tau$ between $\probcancel$ and length variables for the different dataset splits. Values with $^*$ indicate statistical significance ($p < 0.05$).}
\label{tab:kendall_cancel}
\end{table}

\end{document}